%% file: main.tex
\documentclass[runningheads]{llncs}

\usepackage{eccv}

\usepackage{eccvabbrv}

\usepackage{graphicx}
\usepackage{booktabs}

\usepackage[accsupp]{axessibility}  
\usepackage{caption}
\usepackage{wrapfig}
\usepackage{placeins}

\usepackage{pifont}
\usepackage[dvipsnames]{xcolor}
\newcommand{\cmark}{\textcolor{Green}{\ding{51}}}
\newcommand{\xmark}{\textcolor{Red}{\ding{55}}}

\usepackage{hyperref}
\usepackage{xr-hyper}

\usepackage{orcidlink}

\usepackage{multirow}
\usepackage{adjustbox}
\usepackage{makecell}
\usepackage{enumitem}

\begin{document}

\title{Intermediate Text Representation Guided Text-to-Image Generation for Enhancing  One-and-Only Alignment} 
\titlerunning{Intermediate Text Representation Guided One-and-Only Alignment}

\author{Soyoun Won\inst{1}\orcidlink{0009-0002-7107-7411} \and
Aryan Yazdan Parast\inst{1}\orcidlink{0009-0007-2313-7551} \and
Basim Azam\inst{1}\orcidlink{0000-0002-3367-6467} \and 
Jean Honorio\inst{1}\orcidlink{0000-0002-6448-0598} \and 
Naveed Akhtar\inst{1}\orcidlink{0000-0003-3406-673X}}

\authorrunning{S.~Won et al.}

\institute{
The University of Melbourne, Australia\\
\email{\{soyoun.won, aryan.yazdanparast\}@student.unimelb.edu.au}, \\
\email{\{basim.azam, jean.honorio, naveed.akhtar1\}@unimelb.edu.au}
}

\maketitle

\begin{abstract}
Text-to-image (T2I) diffusion models often fail to faithfully render explicit textual descriptions, instead defaulting to strongly learned visual priors due to a phenomenon referred to as \textit{concept association bias}. 
We show that such bias is particularly strong for \textit{one-and-only (OAO)} objects, entities that exist in a single canonical form, such as celestial bodies, landmarks, and artworks. The deeply ingrained visual identity for these concepts often resists modification through prompting alone. 
Addressing this challenge, we first identify through an information-theoretic analysis that the final text embedding discards concept-level information present in the intermediate-layer text representations, reducing the mutual information available to the subsequent denoising process.
We then propose Intermediate Text Representation (IR)-guided diffusion, which injects intermediate hidden states of the text encoder into the conditioning signal during early denoising steps, recovering suppressed concepts without any additional training, optimization, or external models. 
To systematically evaluate the challenging task of aligning generative outputs with unusual prompts for OAO objects, we introduce OAO-AttackBench, a benchmark comprising counterfactual prompts that directly conflict with the core visual identity of OAO objects. Experiments on four benchmarks, including OAO-AttackBench, show that our method achieves up to a 19.1 percentage-point improvement in VQAScore while preserving generation fidelity and human preference. 
Project page: \url{https://soyoun-won.github.io/one-and-only-ir-guidance/}.
\end{abstract}

\section{Introduction}
\label{sec:intro}

\begin{figure}[t]
    \centering
    \includegraphics[width=0.85\linewidth]{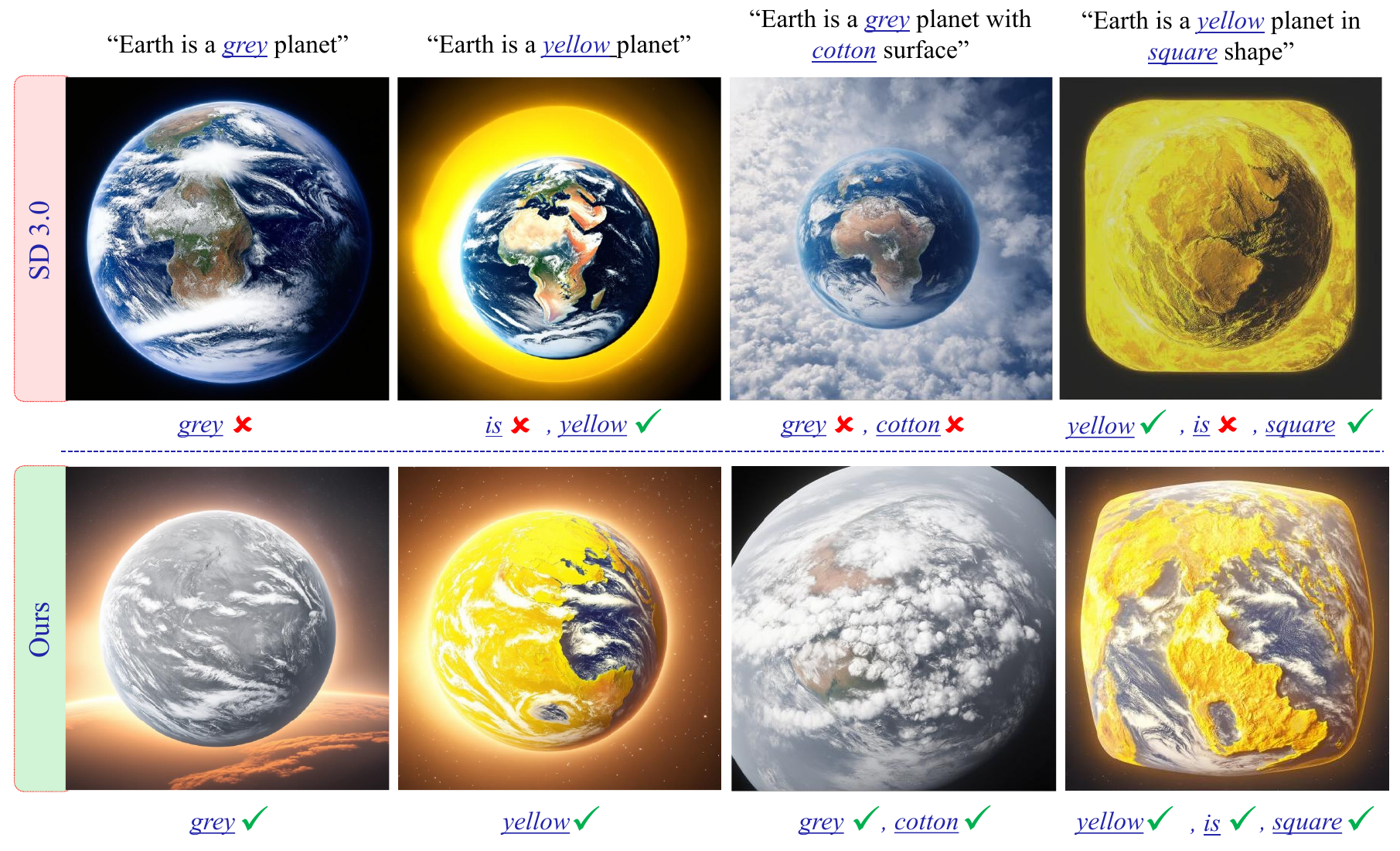}
    \caption{Concept association bias in one-and-only entities. SD3.0 defaults to its learned visual prior for the Earth, ignoring explicitly specified attributes (\xmark). 
    SD3.0 fails to generate individual concepts, and where it does (yellow, square), it fails to appropriately associate  them with the Earth. Our method faithfully reflects all specified concepts and their relationships (\cmark) without additional training or external models. }
    \label{fig:intro}
\end{figure}

Recent text-to-image (T2I) diffusion models have demonstrated remarkable ability to generate photorealistic images from natural language prompts~\cite{rombach2022high,esser2024scaling, islam2026genmix, islam2025context}. Despite this progress, these models frequently fail to faithfully reflect the input prompt, instead defaulting to strongly learned visual priors~\cite{whoops,chefer2023attend}. 
Several methods have been proposed to mitigate this issue~\cite{he2024implicit,orgad2023editing}; however, these approaches require prior knowledge of the specific prompt at inference time, which is difficult to obtain in practice.

Even when a text prompt is underspecified, state-of-the-art T2I models generate high-quality images by filling in the details from the internal knowledge acquired during the training process~\cite{orgad2023editing,yang2024position}. While this ability produces plausible outputs in a typical setting, it becomes a liability when the prompt explicitly contradicts the model's learned associations~\cite{whoops, fu2024commonsense, huberman2025image}. We find that in such cases, T2I models tend to ignore the explicit textual instruction in favour of their ingrained priors.

We refer to this phenomenon as \textit{concept association bias}: the tendency of a model to bind certain visual attributes so tightly to a concept that explicit textual overrides are suppressed. 
Importantly, we do not frame this \textit{bias} as inherently harmful. It is precisely what enables T2I models to generate rich images from sparse descriptions. However, it fundamentally limits the realization of creative or counterfactual prompts. 
We observe that concept association bias is particularly strong for entities with little variation in their visual form and becomes most acute for what we call \textit{one-and-only} entities, entities that exist in a single canonical instance, such as the Earth, the Mona Lisa, or the Eiffel Tower. 
We provide a motivating example in Figure~\ref{fig:intro} using the prompt, \textit{``Earth is a \{color\} planet''}, where the target color attribute deviates from the canonical blue-green appearance of the Earth. Stable Diffusion 3.0 (SD3.0) consistently fails to override its prior, either ignoring the color attribute entirely or failing to associate it with the Earth. 

Our work is motivated by a key observation: simple concepts emerge in the early layers but some of these concepts are lost in the final embedding~\cite{toker2024diffusion}.
We hypothesize that the concepts ignored in images generated by T2I models are often encoded in the intermediate layers of the text encoder, yet fail to be materialized due to excessive abstraction.
We formalize this intuition under an information-theoretic perspective, showing that intermediate hidden states carry greater mutual information with the input text than the final embedding, and that conditioning on these richer representations during the denoising stage of T2I generation can recover the suppressed attribute-level information.  

Based on this analysis, we propose Intermediate Text Representation (IR)-guided diffusion, which injects intermediate representations of text encoders into the conditioning signal during the early denoising steps. IR-guidance operates entirely within the existing model at inference time, requiring no additional training, optimization, or external models.  
Unlike prior training-free approaches that require prior knowledge about text prompts~\cite{chefer2023attend}, our method leverages an untapped internal resource of the text-to-image pipeline itself and employs it for any concept in the prompt without manual specification. 

We are particularly interested in the alignment of generative outputs for one-and-only (OAO) objects. To rigorously evaluate this largely underexplored setting, where concept association bias is especially strong due to the fixed visual identity of OAO objects, we introduce a new benchmark, OAO-AttackBench. OAO-AttackBench is designed to challenge the core visual identity of OAO entities across three categories: celestial objects, landmarks, and artworks.
We evaluate our method on OAO-AttackBench, as well as on existing datasets that challenge input alignment, including  Whoops~\cite{whoops}, Gecko(R), and Gecko(S)~\cite{gecko}, to demonstrate generalization beyond OAO entities. Experimental results confirm that IR-guidance effectively steers the model toward the explicit text prompt, overriding deeply learned associations while preserving generation fidelity and human preference. Our contributions are summarized below:
\begin{itemize}[label = \textbullet]
     \item Through an information-theoretic analysis, we show that conditioning on intermediate text representations with greater mutual information improves text-to-image alignment for prompts with strong learned associations. 
    \item We propose a training- and optimization-free method for improved T2I generation that leverages intermediate text representations during the denoising process to recover suppressed concepts without external models or predefined concept sets. 
    \item We introduce OAO-AttackBench, a benchmark of counterfactual prompts targeting the core visual characteristics of one-and-only (OAO) entities across three categories, providing a systematic evaluation protocol for concept association bias. 
\end{itemize}

\section{Related Work}
Our work draws on three lines of research: compositional T2I generation, bias in T2I diffusion models, and leveraging internal representations of the text encoders. We review each below and position our approach relative to prior methods.

\noindent\textbf{Compositional Text-to-Image Generation.} 
Faithfully composing multiple concepts specified in a text prompt remains a central challenge in text-to-image generation.
Existing lines of work include manipulating the tokens of the text embedding~\cite{hu2024token, seo2025geometrical}, modifying cross-attention layers~\cite{hertz2022prompt, kim2025text, phung2024grounded}, or incorporating large language models~\cite{hu2024ella, sun2025dreamsync}. However, these approaches often require prompt-specific information at inference time, introduce substantial computational overhead due to additional training or optimization procedures, or rely on external models.

\noindent \textbf{Bias in Text-to-Image Diffusion Models.}
Previous works have revealed the biased nature of T2I diffusion models, with extensive work focusing on gender bias~\cite{kim2025rethinking, li2025fair, chuang2023debiasing, friedrich2023fair, azam2025plug, vice2026fairness, parast2025ddb} and cultural bias~\cite{zhang2025joint, shen2023finetuning, um2025minority} demonstrating that certain occupations are more strongly associated with particular genders or races. While widely studied, existing formulations of bias are often limited and require predefined sets. 
In a more general sense, T2I models inherit strong association biases between objects and their typical visual attributes from training data, often neglecting explicitly specified attributes in the prompt~\cite{rassin2024linguistic,huang2023t2icompbench, trusca2024object}. For instance, objects such as tomatoes and sunflowers resist recoloring because their typical attributes dominate the cross-attention maps~\cite{rassin2024linguistic}. 
Editing these associations through projection space has shown promise~\cite{orgad2023editing}, but it requires prior knowledge of which associations to target. Our method does not require any predefined bias set and can address concept associations without prior specification of the target correlations.

Notably, these approaches often aim to remove certain features from generative outputs or introduce diversity associated with the concepts. In other words, they typically aim to address cases where the model associates certain concepts that the user did not explicitly request. 
Our objective is fundamentally different: we address the complementary problem, where the model fails to produce concepts that the user explicitly specified, due to strong association priors ingrained in the model's training data.

\noindent \textbf{Internal Representations of Text Encoders.}
Recent work has shown that intermediate hidden states of text encoders carry information missing from the final text embedding~\cite{toker2024diffusion,li2025does,kim2025text}. Diffusion Lens~\cite{toker2024diffusion} investigated the hidden states by using them as textual conditions for T2I models.
The authors found that simple concepts emerge in the earlier layers, but some of these are absent from the generated images.

Kim et al.~\cite{kim2025text} found that the self-attention map of the text encoder encodes syntax information within the prompt and proposed a method that aligns the cross-attention maps of the denoising network to the self-attention maps. However, they require optimization, which adds computational complexity and only works in diffusion models with a single text encoder.

\section{Proposed Method}
We present IR-guided diffusion, a training-free method that leverages intermediate hidden states of the text encoder to recover concept information lost in the final embedding. We first formalize the concept omission problem through the lens of information theory (\S\ref{sec:problem}), then show that intermediate text representations provably carry richer information (\S\ref{sec:theory}), describe how this signal is injected into the denoising process in our method (\S\ref{sec:injection}), and introduce an adaptive scheduler that automatically determines the stopping point for injection (\S\ref{sec:scheduler}). 

\subsection{Problem Formulation}
\label{sec:problem}
Consider a text prompt $Y$ that describes an object $A$ together with a specified attribute $b$ (e.g., $A =$ ``Earth'' and $b=$ ``purple'' in the prompt ``\textit{Earth is a purple planet}''). We write 
$Y = \{A, b\}$, where both $A$ and $b$ may themselves consist of sub-concepts. Let $x_t$ be the latent variable at diffusion timestep $t$ and $p_\theta(x_t \mid \cdot)$ be the conditional distribution parameterized by $\theta$. A concept omission arises when the model ignores the specified attribute. This is characterized by the scenario where images generated with and without $b$ are distributionally similar. Formally,
\begin{equation}
\begin{aligned}
\label{eq:kl_zero}
D_{\mathrm{KL}}\!\bigl(p_\theta(x_t \mid A, b) || p_\theta(x_t \mid A) \bigr) \approx 0, \\
\end{aligned}
\end{equation}
where $D_{\mathrm{KL}}$ denotes the Kullback--Leibler divergence. From an information-theoretic perspective, Eq.~\eqref{eq:kl_zero} is equivalent to
\begin{equation}
\begin{aligned}
I_\theta(x_t;\, b \mid A) \approx0,
\end{aligned}
\end{equation}
where $I_\theta(\cdot;\cdot\mid\cdot)$ denotes the conditional mutual information under the model distribution parameterized by $\theta$.
This suggests that the generated latent contains negligible information about $b$ once the object $A$ is accounted for. By the chain rule of mutual information, the total information decomposes as 
\begin{equation}
\begin{aligned}
\label{eq:chain_rule}
I(Y;\, x_t) = I(A; \, x_t) + I(b; \, x_t \mid A). \\
\end{aligned}
\end{equation}
Our goal is to increase $I(b;\, x_t \mid A)$, the attribute-specific information in the generated image, while preserving $I(A; \, x_t)$. 
This formulation naturally handles the concept association bias in the generated image. 

\subsection{Theory: Intermediate Representations}
\label{sec:theory}
A typical contemporary text encoder processes the text prompt $Y$ through a sequence of transformer blocks. Let $c_h$ denote the representation at an intermediate block of the underlying network, and $c_f$ denote the representation of the final one. Without loss of generality, we assume a simple network in our analysis that has only these two blocks. Since each processing block is a deterministic function of its input, the encoder forms a Markov chain:
\begin{equation}
\label{eq:markov}
Y \;\rightarrow\; c_h \;\rightarrow \; c_f,
\qquad c_f = h(c_h),
\end{equation}
where $h$ is a deterministic transformation function that maps $c_h$ to $c_f$ for our simple network, and denotes an arbitrary composition of processing blocks for a more complex network.

\begin{lemma}
\label{lem:dpi}
Since $c_f$ is a deterministic function of $c_h$, the chain $Y \rightarrow c_h \rightarrow c_f$ satisfies the Markov property.
By data processing inequality (DPI)~\cite{cover2006elements},
\begin{equation}
    \label{eq:dpi}
    I(Y;\, c_h) \;\ge\; I(Y;\, c_f).
\end{equation}
\end{lemma}
\noindent 
A formal proof of the lemma is provided in Supp. Mat. \S A. Intuitively, a later abstraction block in a network can only transform the signal from an earlier block, which can either preserve or lose information. Eq.~\eqref{eq:dpi} establishes that intermediate hidden states carry at least as much mutual information with the text prompt as the final embeddings. Since the denoising network conditions on the text representation, we posit that richer conditioning can translate to richer latents for the model. Mathematically, 
\begin{equation}
    \label{eq:latent_mi}
    I(Y;\, x_t^h) \;\ge\; I(Y;\, x_t^f),
\end{equation}
where $x_t^h$ and $x_t^f$ denote the latent at timestep $t$ when conditioned on $c_h$ and $c_f$, respectively. We validate this assumption in \S\ref{sec:experiments}: images conditioned on $c_h$ consistently recover the specified attribute that $c_f$ fails to express. 

Prior work has shown that object-level concepts emerge in early layers and are largely preserved through later layers~\cite{toker2024diffusion}, suggesting $I(A;\, x_t^h) \approx I(A;\, x_t^f)$. Combining this with Eq.~\eqref{eq:latent_mi} and the chain rule (Eq.~\eqref{eq:chain_rule}), we get:
\begin{equation}
    \label{eq:attribute_mi}
    I(b;\, x_t^h \mid A) \;\ge\; I(b;\, x_t^f \mid A).
\end{equation}
This implies that conditioning on the intermediate representation can enhance attribute-specific mutual information during denoising. 

\subsection{IR-guided Diffusion}
\label{sec:injection}
\begin{figure}[tb]
  \centering
  \includegraphics[width=\textwidth]{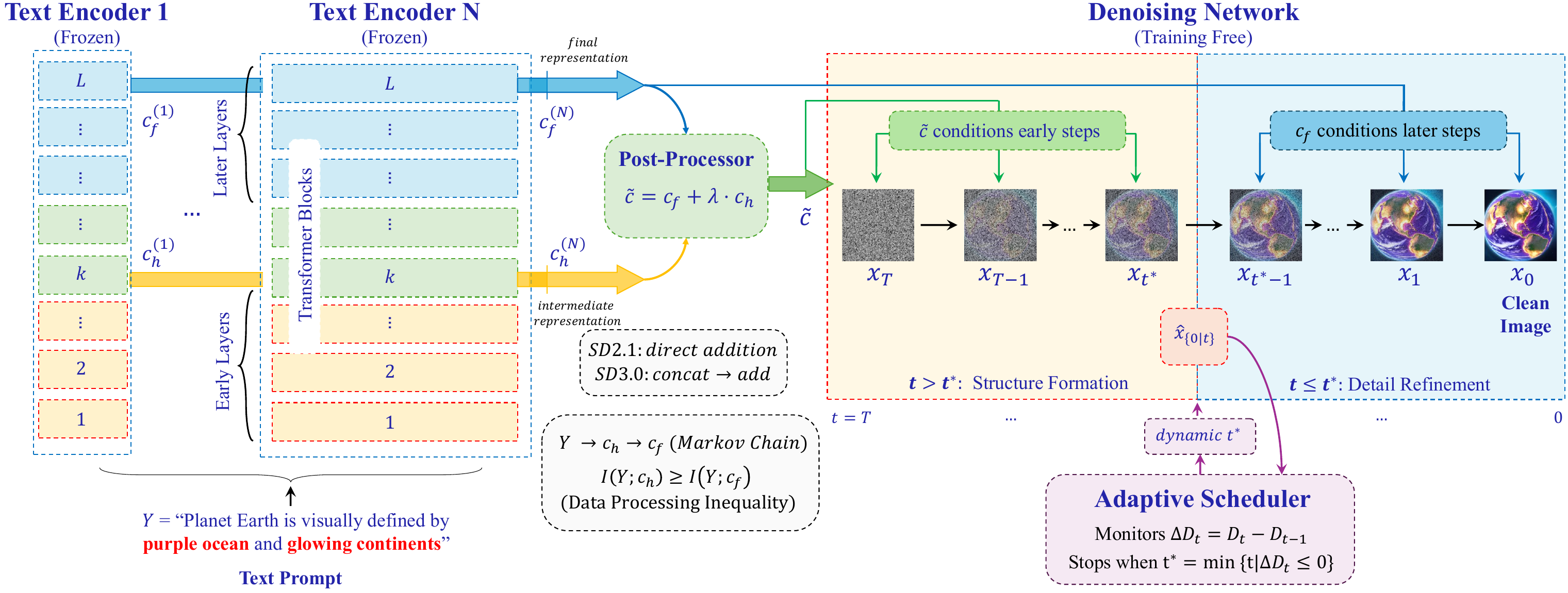}
  \caption{
(Left) The frozen text encoder produces the final text embedding $c_f$ (layer $L$) and an intermediate hidden state $c_h$ (layer $k$), which carries information suppressed in $c_f$.
(Middle)~The Post-Processor constructs an IR-guided embedding $\tilde{c} = c_f + \lambda c_h$, injecting suppressed concept information back into the conditioning signal. (Right)~$\tilde{c}$ conditions the early denoising steps  ($t > t^*$) where global structure is determined, while $c_f$ conditions the later steps ($t \le t^*$) for detail refinement. The Adaptive Scheduler dynamically determines $t^*$ by monitoring structural acceleration $\Delta D_t$. No training or optimization is required.
}
  \label{fig:example}
\end{figure}

Motivated by the analysis above, we inject intermediate representations into the conditioning signal during denoising. The overview is shown in Figure \ref{fig:example}. We treat $c_h$ as a complementary guidance signal and construct an \textit{IR-guided embedding}:
\begin{equation}
    \label{eq:injection}
    \tilde{c} = c_f + \lambda c_h ,
\end{equation}
where $\lambda \in [0,1]$ controls the injection strength. Prior to injection, we apply the text encoder's final layer normalization to $c_h$ to align its distribution with $c_f$; see Supp. Mat. \S B for a distributional analysis. The modified score function is then
\begin{equation}
    \label{eq:score}
    \nabla_{x_t} \log p_\theta(x_t \mid \tilde{c}).
\end{equation}

\noindent \textbf{Multi-Encoder Architectures.} Diffusion architectures with a single text encoder work directly with Eq.~\eqref{eq:injection}. For architectures employing $N$ text encoders  whose outputs are concatenated before conditioning, we extract intermediate representations from each encoder independently, concatenate them following the native scheme, and  add the result to the concatenated final embedding:
\begin{equation}
    \label{eq:mutlti_encoder}
    \tilde{c}
    = \bigl[\, c_f^{(1)} \;\|\; \cdots \;\|\; c_f^{(N)} \,\bigr]
    + \lambda \, \bigl[ \, c_h^{(1)} \;\|\; \cdots \;\|\; c_h^{(N)} \bigr],
\end{equation}
where $[\cdots \| \cdots]$ denotes the concatenation and superscripts index the encoders. This preserves the native conditioning interface for each architecture without requiring any architectural modification. 

\noindent \textbf{Post-Processing Strategies.}
To validate the effectiveness of additive injection as the post-processing strategy, we formulate four alternatives. Following~\cite{lemon}, the final layer normalization is applied to the intermediate representation for all variants. 
\textit{None} uses the intermediate representation directly $\tilde{c} =c_h$. \textit{Mean-Std} normalizes $c_h$ to match the samplewise statistics of $c_f$:
\begin{equation}
\tilde{c}_\text{mean-std}
= \frac{c_h - \mu(c_h)}
    {\sigma(c_h)}
    \cdot 
    \sigma(c_f) + \mu(c_f),
\end{equation}
where $\mu(\cdot)$ and $\sigma(\cdot)$ denote the mean and standard deviation, respectively. \textit{Orthogonal projection} decomposes $c_h$ into a component aligned with $c_f$ and a residual. The projection matrix is
%
\begin{equation}
P = c_f\,(c_f^\top c_f )^{-1}\,c_f^\top,
\end{equation}
yielding two variants: \textit{Projection} ($\tilde{c}_{\text{proj}} =P c_h$) and \textit{Residual} ({$\tilde{c}_{\text{res}} = c_h - P c_h$}). See Supp. Mat. \S B for a comparative analysis.

\subsection{Adaptive Scheduling}
\label{sec:scheduler}
While intermediate representations carry richer information, conditioning on $\tilde{c}$ throughout the entire denoising process can introduce artifacts.
Since diffusion models establish global structure in early denoising steps and refine fine-grained detail in later steps\cite{wang2023diffusion, chefer2023attend, hertz2022prompt}, we condition on $\tilde{c}$ only during the early, structure-forming phase and revert to the standard embedding $c_f$ once the layout stabilizes.  

To determine the transition point $t^*$ dynamically, we monitor structural changes in the estimated clean image $\hat{x}_0$ via Tweedie's formula~\cite{efron2011tweedie,robbins1992empirical} at each timestep: 
\begin{equation}
    \label{eq:tweedie}
    \hat{x}_{0|t}
    = \frac{x_t - \sqrt{1-\bar{\alpha}_t}\,
    \epsilon_\theta(x_t,t)}{\sqrt{\bar{\alpha}_t}},
\end{equation}
where $\bar{\alpha}_t$ is a set of monotonically decreasing parameters that control the noise level and $\epsilon_\theta$ denotes the predicted noise parameterized by $\theta$. Then, we measure the magnitude of structural updates as the squared difference between consecutive predictions
\begin{equation}
    \label{eq:structural_update}
    D_t = \bigl\|\, \hat{x}_{0|t} - \hat{x}_{0|t-1} \,\bigr\|_2^2,
\end{equation}
and define the \textit{structural acceleration} as its first-order difference:
\begin{equation}
    \label{eq:accelration}
    \Delta D_t = D_t - D_{t-1}.
\end{equation}
$\Delta D_t > 0$ indicates that the structural changes are intensifying, and $\Delta D_t \le 0$ suggests that the layout is stabilizing. The transition point is the first timestep at which acceleration becomes non-positive:
\begin{equation}
    \label{eq:stopping}
    t^* = \min\, \bigl\{\, t \;\big|\ \Delta D_t \le 0 \,\bigr\}.
\end{equation}
The complete denoising process is:
\begin{equation}
    \label{eq:full_process}
    x_{t-1} = 
    \begin{cases}
        \text{denoise}(x_t,\, \tilde{c}) & \text{if } t > t^*, \\[2pt]
        \text{denoise}(x_t, \, c_f) & \text{if } t \le t^*.
    \end{cases}
\end{equation}
This adaptive strategy avoids the need for a fixed hyperparameter to control the transition, and the overhead of computing $D_t$ is negligible compared to the denoising step itself; see Supp. Mat. \S K for computational cost analysis.

\subsection{OAO-AttackBench}
Models pick up biases from the datasets they are exposed to during training~\cite{yang2024position, seshadri2024bias, carlini2023extracting}. We observe from the previous section that such concept association bias is severe for one-and-only entities, objects that exist in a single canonical form.
To address the challenging cases of concept association, we present \textbf{OAO-AttackBench}. 
OAO-AttackBench is designed to attack the implicit knowledge embedded in one-and-only entities. The term \textit{attack} reflects the dataset's nature, which contradicts or violates the core visual identity of the one-and-only entities. 
Unlike everyday objects, one-and-only entities are strongly tied to a single visual form, making generation from unusual descriptions more challenging.
OAO-AttackBench consists of three categories: celestial objects, landmarks, and artworks, each of which is sub-categorized by shape and material, pattern and material, and genre, style and material, for a total of 504 prompts. Details on the dataset are provided in Supp. Mat. \S C.

\vspace{-1em}
\section{Experiments}
We evaluate IR-guided diffusion on four benchmarks spanning counterfactual, commonsense-violating, and general-purpose compositional prompts across multiple diffusion backbones. Our experiments address three questions: (i)~does IR-guidance improve text-to-image alignment for prompts that trigger concept association bias? (ii)~does it preserve generation fidelity and human preference? (iii)~does it generalize to standard prompts without degrading quality?

\label{sec:experiments}
\vspace{-1em}
\subsection{Experimental Setup}
\noindent \textbf{Datasets.}
We evaluate our method on four benchmarks: our proposed OAO-AttackBench, Whoops~\cite{whoops}, Gecko(R), and Gecko(S)~\cite{gecko}.
\textit{OAO-AttackBench} is designed to evaluate a challenging form of concept association bias, where each subject is assumed to exist in only one canonical form. 
\textit{Whoops} includes ``weird'' prompts that are designed to violate commonsense about everyday objects, testing whether models can override typical associations such as physical rules, social knowledge, and cultural norms. \textit{Gecko} is a dataset containing 2K prompts, divided into Gecko(R) and Gecko(S). \textit{Gecko(R)} contains texts resampled from existing datasets
\footnote{TIFA~\cite{tifa}, Stanford Paragraph\cite{stanford}, Localized Narratives\cite{localized_narrative}, CountBench\cite{countbench}, VRD\cite{vrd}, DiffusionDB\cite{diffusiondb}, MJ\cite{mj}, PoseScript\cite{posescript}, Whoops\cite{whoops}, and DrawText-Creative\cite{drawtext}}, and
\textit{Gecko(S)} contains LLM-generated texts that span categories covered less in Gecko(R). 

\noindent \textbf{Models.}
We implement IR-guided diffusion on Stable Diffusion (SD)~2.1~\cite{rombach2022high} and~3.0~\cite{esser2024scaling}. SD2.1 employs a pretrained OpenCLIP ViT-H/14 text encoder, and SD3.0 has three pretrained text encoders: OpenCLIP ViT-G, CLIP ViT-L, and T5-XXL. 
Additional implementation details are provided in Supp. Mat. \S D.
To exploit word-level distinctiveness suppressed in the final embedding but retained in earlier-layer representations, we extract intermediate representations after the first transformer block of the text encoder ($k\!=\!1$). Analyses across encoder depth are provided in Supp. Mat. \S E.

\noindent \textbf{Metrics.} 
We measure text-to-image alignment with two complementary metrics. \textit{CLIPScore}~\cite{clipscore} computes the cosine similarity between image and text embeddings from OpenAI CLIP ViT-L/14. \textit{VQAScore}~\cite{VQAScore} evaluates compositional alignment by querying the CLIP-FlanT5 VQA model with the question ``Does this figure show \{text prompt\}?'' and averaging the probability of ``yes''. Because VQAScore probes the presence of fine-grained attributes, it is more sensitive to concept omission than CLIPScore, which captures global semantic similarity. To verify IR-guidance does not introduce artifacts, we report the \textit{Kernel Inception Distance (KID)}~\cite{kernel} between the generated images and COCO~\cite{coco} reference images. Human preference is quantified with \textit{HPSv2} and \textit{HPSv2.1}~\cite{hps}.

\subsection{Results}
\label{sec:results}

\begin{figure}[t]
    \centering
    \includegraphics[width=0.95\linewidth]{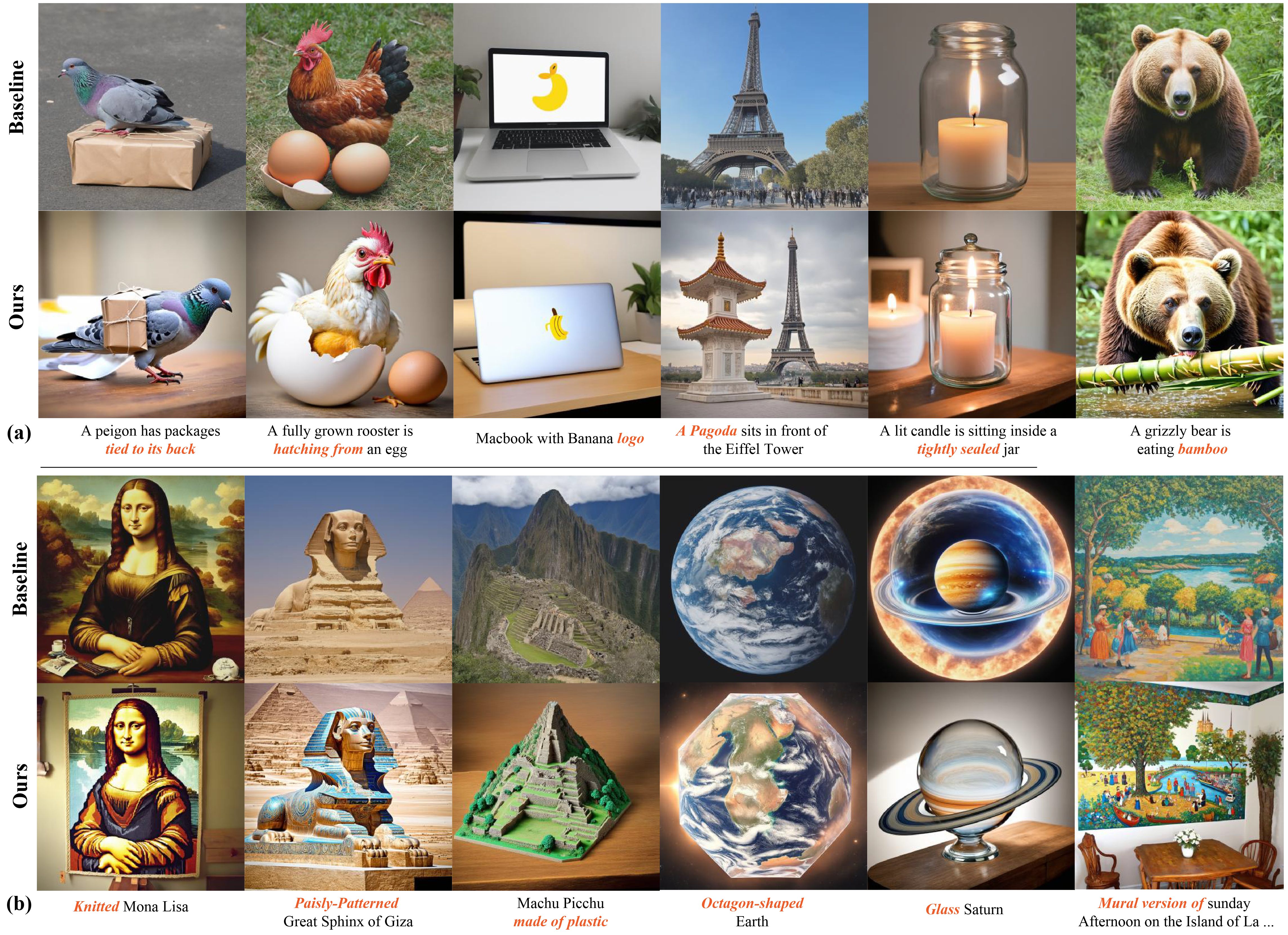}
    \caption{\textit{Qualitative comparison on SD3.0}. (a) Whoops prompts targeting everyday objects. The baseline generates the individual concepts but fails to express the specified attribute (\textcolor{orange}{orange text}). (b) OAO-AttackBench prompts targeting one-and-only entities. The baseline preserves canonical appearances despite explicit contradictions in the prompt; the Mona Lisa remains a painting, the Earth remains spherical, and Saturn remains opaque. IR-guided diffusion successfully overrides these priors (\textit{Knitted}, \textit{Paisley Patterned}, \textit{made of plastic}, \textit{Octagon-shaped}, \textit{Glass}) while preserving object identity.  }
    \label{fig:samples-sd3}
    \vspace{-1em}
\end{figure}

\noindent \textbf{Qualitative Comparison.} Figure~\ref{fig:samples-sd3} presents side-by-side comparisons on SD3.0. On Whoops (top), the baseline generates both individual concepts but fails to express their specified relationships: the packages appear detached from the pigeon rather than \textit{tied to its back}, and the candle sits in an open jar rather than \textit{tightly sealed}. IR-guidance recovers these suppressed attributes and relationships. The contrast is more striking on OAO-AttackBench (bottom), where the baseline preserves canonical appearances despite explicit contradictions in the prompt. Saturn remains opaque despite ``\textit{Glass} Saturn''; Earth remains spherical despite ``\textit{Octagon-shaped} Earth''; the Mona Lisa retains its painted appearance despite ``\textit{Knitted} Mona Lisa.'' IR-guidance successfully overrides these deeply learned priors while preserving object identity. More qualitative results are provided in Supp. Mat. \S F.

\begin{table*}[!t]
\centering
\caption{Quantitative comparison on Whoops and OAO-AttackBench. \textbf{Bold} and \text{italic} values denote the best and the second-best scores, respectively. 
}
\label{tab:result_aoa}
\vspace{-0.2cm}
\setlength{\tabcolsep}{6pt}
\begin{adjustbox}{max width=\textwidth}

\begin{tabular}{c | c | l | cc | cc | ccc | c}
\toprule
\multicolumn{3}{c|}{} 
& \multicolumn{7}{c|}{\textbf{OAO-AttackBench}} 
& \multirow{4}{*}{\textbf{Whoops}} \\
\cmidrule(lr){4-10}
\multicolumn{3}{c|}{} 
& \multicolumn{2}{c|}{Celestial Objects} 
& \multicolumn{2}{c|}{Landmarks} 
& \multicolumn{3}{c|}{Artworks} 
&  \\
\cmidrule(lr){4-5} \cmidrule(lr){6-7} \cmidrule(lr){8-10}
\multicolumn{3}{c|}{} 
& Shape & Material 
& Pattern & Material 
& Genre & Style & Material 
&  \\
\midrule

\multirow{12}{*}{\rotatebox[origin=c]{90}{\textbf{VQA}}}
& \multirow{6}{*}{SD3.0}
& Baseline  & 0.514 & 0.687 & 0.778 & 0.525 & 0.719 & \textit{0.733} & 0.657 & 0.778 \\
&  & None      & 0.509 & 0.716 & 0.773 & 0.630 & 0.750 & 0.693 & 0.706 & 0.773 \\
&  & Projection      & \textit{0.563} & 0.703 & \textit{0.792} & \textit{0.658} & 0.748 & 0.733 & 0.709 & \textit{0.792} \\
&  & Residual & 0.516 & 0.694 & 0.779 & 0.630 & \textbf{0.778} & 0.700 & \textit{0.726} & 0.779 \\
&  & Mean-Std & 0.544 & \textit{0.725} & 0.760 & 0.617 & 0.734 & 0.701 & 0.716 & 0.760 \\
&  & Ours      & \textbf{0.565} & \textbf{0.726} & \textbf{0.802} & \textbf{0.716} & \textit{0.753} & \textbf{0.736} & \textbf{0.756} & \textbf{0.802} \\

\cmidrule(lr){2-11}

& \multirow{6}{*}{SD2.1}
& Baseline  & 0.478 & 0.685 & 0.629 & 0.516 & \textit{0.688} & 0.563 & 0.675 & 0.629 \\
&  & None      & 0.501 & 0.700 & \textit{0.632} & \textit{0.527} & 0.658 & 0.581 & 0.702 & \textit{0.632} \\
&  & Projection      & \textbf{0.522} & 0.701 & 0.627 & 0.511 & 0.646 & 0.568 & \textit{0.702} & 0.627 \\
&  & Residual & \textit{0.509} & \textbf{0.711} & 0.617 & 0.510 & 0.640 & \textit{0.585} & 0.692 & 0.617 \\
&  & Mean-Std & 0.465 & 0.680 & 0.617 & \textbf{0.534} & 0.668 & 0.570 & 0.674 & 0.617 \\
&  & Ours      & \textit{0.509} & \textit{0.703} & \textbf{0.641} & 0.514 & \textbf{0.693} & \textbf{0.622} & \textbf{0.715} & \textbf{0.641} \\

\midrule

\multirow{12}{*}{\rotatebox[origin=c]{90}{\textbf{CLIP}}}
& \multirow{6}{*}{SD3.0}
& Baseline  & 0.257 & 0.244 & 0.291 & 0.263 & 0.263 & 0.278 & 0.263 & 0.291 \\
&  & None      & 0.246 & 0.240 & 0.293 & 0.278 & 0.264 & 0.267 & 0.270 & 0.293 \\
&  & Projection      & \textit{0.258} & 0.245 & \textit{0.293} & \textit{0.284} & 0.269 & \textit{0.279} & 0.271 & \textit{0.293} \\
&  & Residual & 0.249 & 0.244 & 0.291 & 0.276 & \textit{0.269} & 0.276 & \textit{0.272} & 0.291 \\
&  & Mean-Std & 0.252 & \textbf{0.253} & 0.291 & 0.279 & 0.260 & 0.274 & 0.265 & 0.291 \\
&  & Ours      & \textbf{0.259} & \textit{0.253} & \textbf{0.296} & \textbf{0.289} & \textbf{0.274} & \textbf{0.281} & \textbf{0.276} & \textbf{0.296} \\

\cmidrule(lr){2-11}

& \multirow{6}{*}{SD2.1}
& Baseline  & \textit{0.246} & \textbf{0.241} & 0.280 & 0.265 & 0.274 & 0.271 & \textit{0.282} & 0.280 \\
&  & None      & 0.243 & 0.239 & \textbf{0.282} & \textit{0.268} & \textit{0.276} & \textit{0.276} & 0.281 & \textbf{0.282} \\
&  & Projection      & 0.244 & 0.238 & 0.280 & 0.266 & 0.273 & 0.275 & \textbf{0.284} & 0.280 \\
&  & Residual & \textbf{0.247} & \textit{0.239} & 0.279 & 0.265 & 0.273 & 0.276 & 0.282 & 0.279 \\
&  & Mean-Std & 0.245 & 0.232 & 0.278 & \textbf{0.273} & \textbf{0.277} & 0.275 & 0.281 & 0.278 \\
&  & Ours      & 0.239 & 0.239 & \textit{0.281} & 0.266 & 0.272 & \textbf{0.277} & 0.280 & \textit{0.281} \\

\bottomrule
\end{tabular}
\end{adjustbox}
\end{table*}

\begin{figure}[t]
  \centering
  \includegraphics[width=0.97\textwidth]{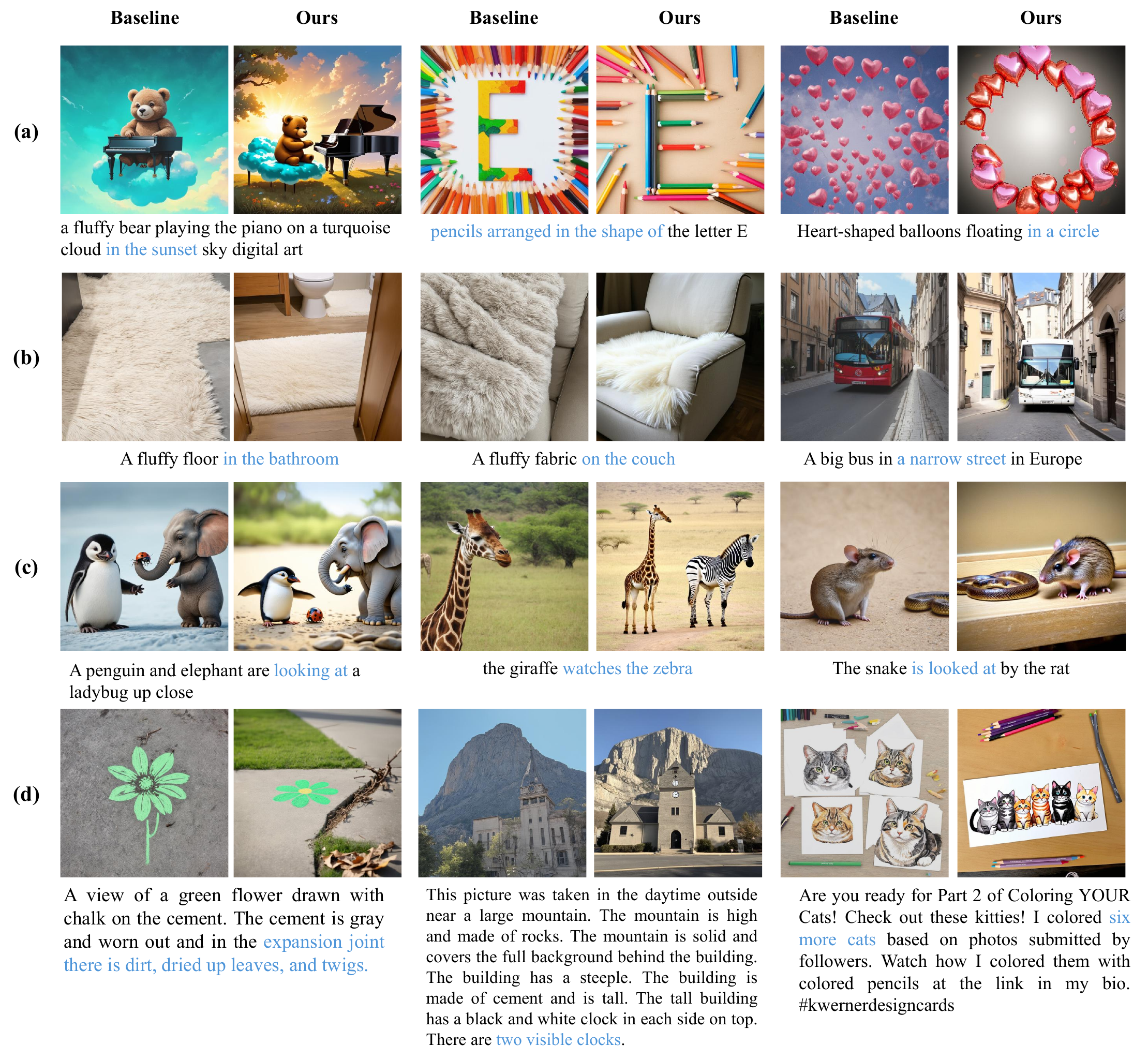}
  \vspace{-1em}
    \caption{Sample comparison on Gecko. \textcolor{RoyalBlue}{Blue text} denotes concepts ignored or misinterpreted in the baseline output. IR-guidance improves (a) general concept omission problem beyond one-and-only entities, (b) environmental context, (c) relational scenarios involving multiple objects, and (d) long-text prompts.}  \label{fig:gecko_all}
\end{figure}

\noindent \textbf{Text-image Alignment.} Table \ref{tab:result_aoa} reports VQAScore and CLIPScore on OAO-AttackBench and Whoops. IR-guidance consistently outperforms baseline under VQAScore, achieving up to a 19.1 percentage-point improvement on OAO-AttackBench. The modest variation in CLIPScore suggests that the global semantic alignment is largely preserved, whereas the improvement in VQAScore indicates enhanced fine-grained text-to-image alignment.
We also compare against the four alternative post-processing strategies defined in \S\ref{sec:injection}: None, Mean-Std, Projection, and Residual. As shown in Table~\ref{tab:result_aoa}, Ours (additive injection) 
consistently outperforms or matches all alternatives. Evaluation results on additional metrics are provided in Supp. Mat. \S I.

\begin{wraptable}{r}{0.53\textwidth}
\vspace{-3.4em}
\caption{KID evaluation results on OAO-AttackBench (lower is better). Our method achieves comparable scores across categories.}
\vspace{0.5em}
\label{tab:results_kid}
\setlength{\tabcolsep}{3pt}
\centering
\resizebox{0.7\linewidth}{!}{
\begin{tabular}{c | c c c}
  \toprule
  & \makecell{Celestial\\Objects}
  & \makecell{Landmarks}
  & \makecell{Artworks} \\
  \midrule
  SD2.1 & 0.0551 & 0.0559 & 0.0624 \\
  \textbf{Ours$_{2.1}$} & \textbf{0.0491} & \textbf{0.0557} & \textbf{0.0616} \\
  \midrule
  SD3.0 & 0.0576 & 0.0583 & \textbf{0.0382} \\
  \textbf{Ours$_{3.0}$} & \textbf{0.0483} & \textbf{0.0500} & 0.0397 \\
  \bottomrule
\end{tabular}
}
\vspace{-2em}
\end{wraptable}

We additionally report Kernel Inception Distance (KID) in
Table~\ref{tab:results_kid}. IR-guided diffusion achieves comparable or better scores than the baseline across all categories, confirming that the method preserves distributional fidelity. Experimental results suggest that IR-guidance not only improves compositional accuracy but also maintains, and in most cases enhances, the overall distribution quality of the generated images.

\noindent \textbf{Human Preference.} Table~\ref{table:human-preference} reports HPSv2 and HPSv2.1 on SD3.0. IR-guided diffusion improves human preference scores on both OAO-AttackBench and Whoops relative to the baseline. The improvement confirms that IR-guided images are not only more faithful to the text prompt but also more visually plausible to human observers.

\noindent \textbf{Generation Beyond Unusual Prompts.} So far, we have examined IR-guidance on the set of prompts that attack the core visual characteristics of one-and-only entities (OAO-AttackBench) and on prompts that violate commonsense (Whoops). Here, we further assess its compositional ability in more general settings.
Table \ref{tab:results_gecko} summarizes the evaluation results on Gecko with SD3.0. As shown in the table, the IR-guided model achieves comparable performance to the baseline. While improvements are modest, the results indicate that IR-guidance does not compromise standard generation quality on relatively general text prompts. 

We visualize samples of IR-guided diffusion on Gecko in Figure \ref{fig:gecko_all}. 
\textit{Concept Omission}: For underrepresented concepts, Figure \ref{fig:gecko_all}(a) demonstrates that IR-guidance effectively mitigates concept omission beyond one-and-only entities.
\textit{Contextual and Relational Alignment}: Figure \ref{fig:gecko_all}(b-c) shows that the baseline generates plausible subjects but often misses the surrounding context or misinterprets interactions between objects. IR-guidance improves contextual cues and object relations.
\textit{Long Prompts}: As shown in Figure \ref{fig:gecko_all}(d), IR-guidance preserves concepts specified in extended textual descriptions while also mitigating concept omission.
Across these diverse settings, IR-guidance improves text-to-image alignment while preserving overall image quality.
\vspace{1mm}

\vspace{-0.5em}
\subsection{Additional Analysis}
\label{sec:analysis}

\begin{wrapfigure}{l}{0.45\textwidth}
  \centering
  \vspace{-2em}
  \includegraphics[width=0.43\textwidth,keepaspectratio]{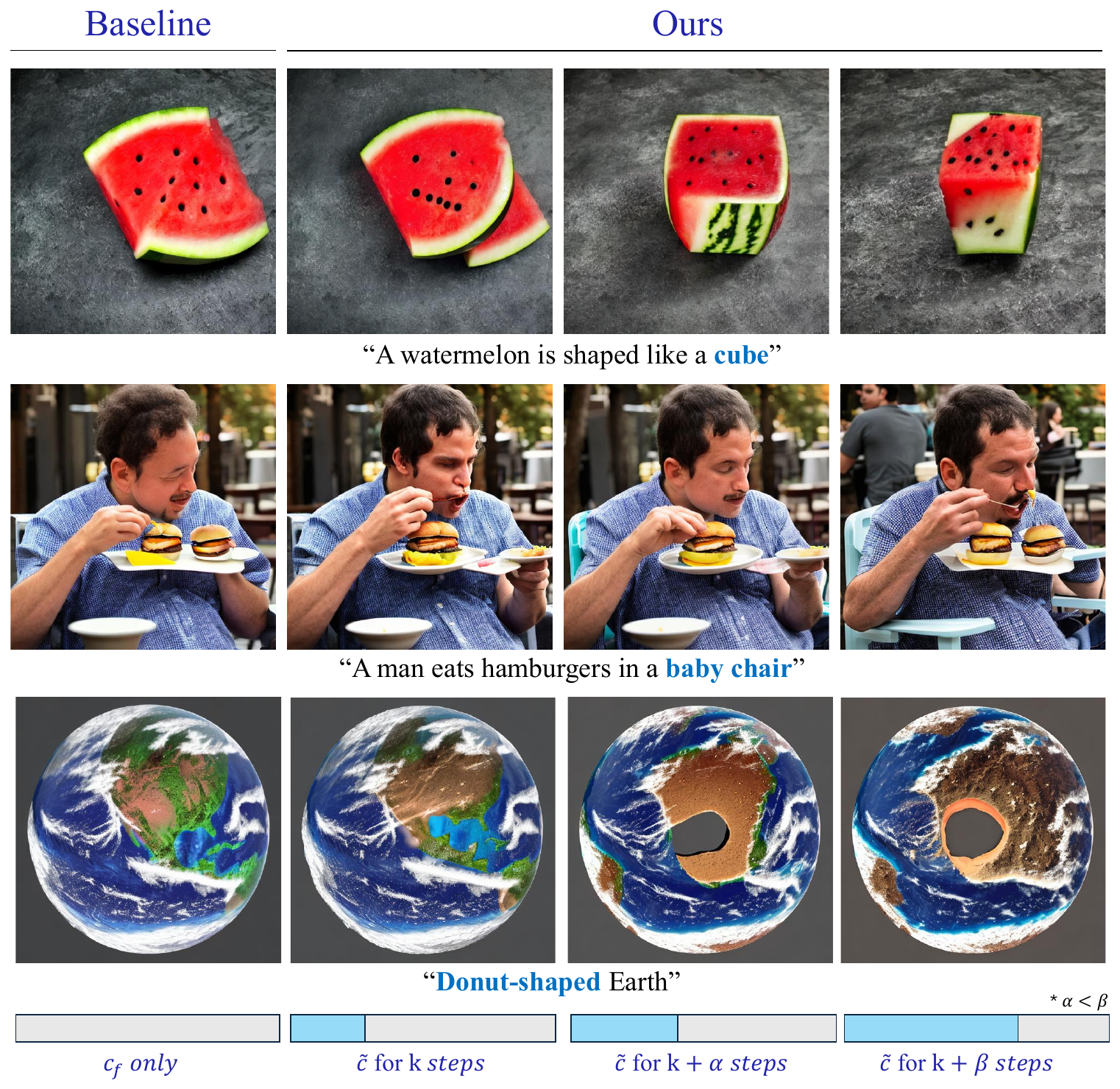}
  \caption{Effect of IR-guidance duration. Each row shows the baseline ($c_f$ only) and three variants where $\tilde{c}$ is used as conditioning for progressively more denoising steps. The timeline bars at the bottom indicate the conditioning schedule. As IR-guidance is extended to more timesteps, the specified attribute (\textcolor{RoyalBlue}{blue text}) is more faithfully expressed.}
  \label{fig:evolve}
  \vspace{-2em}
\end{wrapfigure}

\noindent \textbf{Effect of IR-guidance Duration.}
Figure \ref{fig:evolve} illustrates how the generated images change as IR-guidance is applied over different ranges of timesteps on SD2.1. The first column shows the baseline without IR-guidance, and the subsequent columns progressively extend the range over which IR-guidance is applied.
As IR-guidance is applied over more timesteps, the generated images increasingly exhibit the previously missing concepts.

\begin{figure}[t]
  \centering
  \includegraphics[width=\textwidth]{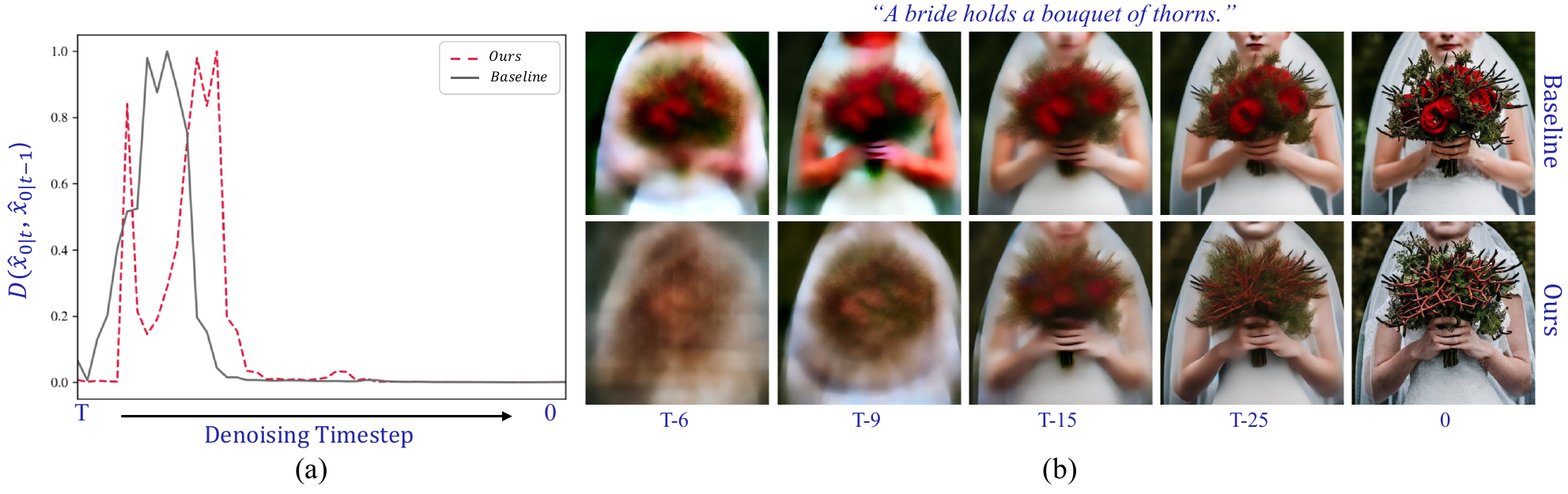}
  \caption{(a) Mean squared difference of predicted images between two consecutive denoising timesteps. (b) Predicted generated image at each denoising timestep on the baseline (top) and IR-guided diffusion (bottom). IR-guided diffusion delays premature structural convergence, allowing concept exploration during early denoising steps.}  
  \label{fig:mse_graph}
  \vspace{-1.5em}
\end{figure}

\noindent \textbf{What does IR-guidance do?} To further investigate the step-wise influence of IR-guidance, we calculate the mean squared difference between predicted samples $\hat{x}_0$ from two consecutive timesteps. Figure \ref{fig:mse_graph}(a) shows the mean squared difference over the denoising process, and Figure \ref{fig:mse_graph}(b) visualizes the predicted samples at various timesteps for SD2.1. A high mean squared difference indicates a substantial change in the predicted image, suggesting a shift in the overall structure of the generated image. 

Figure \ref{fig:mse_graph}(b) shows that the baseline's estimated output already closely resembles the final generated image at the early timestep $T-6$.
In contrast, IR-guided estimation at the same timestep shows a noisier predicted image, where the overall structure is substantially different from the final generated image.
IR-guidance incorporates intermediate text representations for the first few steps, then switches the textual conditioning to the final text embedding. Naturally, this introduces a steep increase in the mean squared difference at the transition point. Interestingly, IR-guided diffusion (dotted red line in Figure \ref{fig:mse_graph}(a)) shows two major local maxima after the transition, suggesting a shift in the denoising trajectory that prevents premature structural convergence.

\begin{table}[t]
\centering
\begin{minipage}[h]{0.4\linewidth}
\centering
\scriptsize
\caption{Evaluation results on human preference metrics.}
\label{table:human-preference}
\setlength{\tabcolsep}{4pt}
\resizebox{\linewidth}{!}{
\begin{tabular}{clcc}
\toprule
Dataset & Model & HPSv2 & HPSv2.1 \\
\midrule
\multirow{2}{*}{\makecell[c]{OAO-\\AttackBench}}
& Baseline & 0.269 & 0.263 \\
& Ours     & \textbf{0.273} & \textbf{0.271} \\
\midrule
\multirow{2}{*}{Whoops}
& Baseline & 0.291 & 0.295 \\
& Ours     & \textbf{0.293} & \textbf{0.299} \\
\bottomrule
\end{tabular}
}
\end{minipage}
\hfill
\begin{minipage}[h]{0.47\linewidth}
\centering
\scriptsize
\caption{Evaluation results on general datasets Gecko(R) and Gecko(S).}
\label{tab:results_gecko}
\setlength{\tabcolsep}{3pt}
\resizebox{\linewidth}{!}{
\begin{tabular}{llcccc}
\toprule
Dataset & Model & CLIP & VQA & HPSv2 & HPSv2.1 \\
\midrule
\multirow{2}{*}{Gecko(R)}
& Baseline & 0.281 & 0.866 & 0.281 & 0.264 \\
& Ours     & \textbf{0.284} & \textbf{0.871} & \textbf{0.284} & \textbf{0.271} \\
\midrule
\multirow{2}{*}{Gecko(S)}
& Baseline & 0.282 & 0.744 & 0.283 & 0.274 \\
& Ours     & \textbf{0.285} & \textbf{0.748} & \textbf{0.286} & \textbf{0.278} \\
\bottomrule
\end{tabular}
}
\end{minipage}
\end{table}

\begin{wrapfigure}{r}{0.52\textwidth}
\vspace{-2em}
  \centering
  \includegraphics[width=0.52\textwidth]{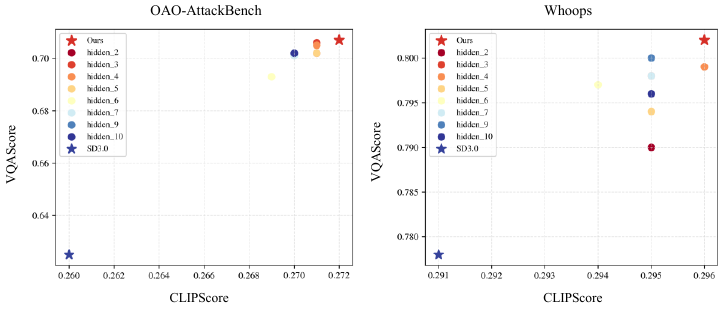}
  \caption{VQAScore vs.\ CLIPScore of $c_h$ from different transformer blocks. Earlier layers yield a higher VQAScore on OAO-AttackBench.
  }
  \label{fig:hidden_X}
  \vspace{-2em}
\end{wrapfigure}

\noindent \textbf{Ablation Study.} Figure \ref{fig:hidden_X} reports VQAScore and CLIPScore across different layers of the text encoder for extracting $c_h$ on SD3.0. 
Earlier layers yield higher VQAScore on OAO-AttackBench, consistent with the observation that word-level information is present in early layers and is compressed in later layers (Supp. Mat. \S E). 
On Whoops, the sensitivity is lower, reflecting the less rigid priors associated with everyday settings. Additional ablation studies are provided in Supp. Mat. \S G.

\section{Conclusion}
In this work, we introduce IR-guided diffusion, which leverages the internal states of the text encoder during the denoising process for prompt-faithful generation, accentuating underrepresented concepts. Our approach requires no predefined set of biases, additional training, optimization, or external models. To validate its effectiveness, we introduce OAO-AttackBench, a challenging dataset of prompts that attack the canonical identity of one-and-only entities. Experimental results show that our approach is effective not only for one-and-only entities but also for everyday objects, while maintaining comparable performance on fidelity and human preference scores compared with the state-of-the-art baseline. 

\section*{Acknowledgements}
Naveed Akhtar is a recipient of the Australian Research Council Discovery Early Career Researcher Award (project \# DE230101058) funded by the Australian Government. This research was supported by The University of Melbourne’s Research Computing Services and the Petascale Campus Initiative.This work
is also partially supported by the Google Research Scholar Program Award.

\bibliographystyle{splncs04}
\bibliography{main,supp}

\clearpage
\appendix
\renewcommand{\thefigure}{S\arabic{figure}}
\renewcommand{\thetable}{S\arabic{table}}
\renewcommand{\theequation}{S\arabic{equation}}
\setcounter{figure}{0}
\setcounter{table}{0}
\setcounter{equation}{0}

\input{supp}

\end{document}

%% file: supp.tex
\title{
Intermediate Text Representation Guided Text-to-Image Generation for Enhancing One-and-Only Alignment \\
(Supplementary Material)
} 

\titlerunning{Intermediate Text Representation Guided One-and-Only Alignment}

\author{Soyoun Won\inst{1}\orcidlink{0009-0002-7107-7411} \and
Aryan Yazdan Parast\inst{1}\orcidlink{0009-0007-2313-7551} \and
Basim Azam\inst{1}\orcidlink{0000-0002-3367-6467} \and 
Jean Honorio\inst{1}\orcidlink{0000-0002-6448-0598} \and 
Naveed Akhtar\inst{1}\orcidlink{0000-0003-3406-673X}}

\authorrunning{S.~Won et al.}

\institute{
The University of Melbourne, Australia\\
\email{\{soyoun.won, aryan.yazdanparast\}@student.unimelb.edu.au}, \\
\email{\{basim.azam, jean.honorio, naveed.akhtar1\}@unimelb.edu.au}
}

\maketitle

\renewcommand{\thesection}{\Alph{section}}
\section{Complete Proof of Lemma 1}
\label{supp_sec:proof}

\begin{lemma}
\label{supp_lem:dpi}
Since $c_f$ is a deterministic function of $c_h$, the chain $Y \rightarrow c_h \rightarrow c_f$ satisfies the Markov properties.
By data processing inequality (DPI)~\cite{cover2006elements},
\begin{equation}
    \label{supp_eq:dpi}
    I(Y;\, c_h) \;\ge\; I(Y;\, c_f).
\end{equation}
\end{lemma}

Recall that $Y$ is the text prompt, $c_h$ and $c_f$ are intermediate and final transformation block representations respectively, and $h$ is a deterministic transformation function that maps $c_h$ to $c_f$, giving the Markov chain $Y \rightarrow c_h \rightarrow c_f$. 
Under the Markov structure, the following conditional independence holds:
\begin{equation}
Y \bot c_f \mid c_h.
\end{equation}
Equivalently, 
\begin{equation}
\label{supp_eq:cond_orth}
I(Y;\, c_f \mid c_h)=0.
\end{equation}
From the chain rule, 
\begin{equation}
\begin{aligned}
I(Y; (c_h, c_f)) = I(Y; c_f) + I(Y; c_h \mid c_f) \\
= I(Y; c_h) + I(Y; c_f \mid c_h).
\end{aligned}
\end{equation}
Substituting Equation~\ref{supp_eq:cond_orth} into the expression above yields
\begin{equation}
I(Y; c_f) + I(Y; c_h \mid c_f) = I(Y; c_h).
\end{equation}
Since mutual information is always non-negative, we obtain
\begin{equation}
I(Y; c_h) \ge I(Y; c_f), 
\end{equation}
suggesting that the mutual information between the input text and the intermediate representation of the text encoder is greater than or equal to that of the input text and the final text representation. 

\section{Distribution Analysis}
\label{supp_sec:dist_analysis}

\begin{figure}[t]
\includegraphics[width=\linewidth]{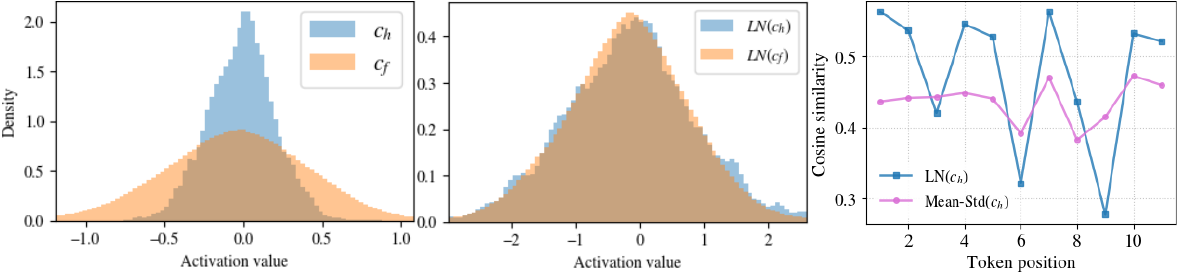}
\caption{ (Left) Distributional mismatch between $c_h$ and $c_f$ (WD:0.2). (Middle) $LN$ reduces WD to 0.05.
(Right) $Mean\text{-}Std(c_h)$ flattens token-wise similarity while $LN$ preserves it. }
\label{supp_fig:dist_analysis}
\end{figure}

The idea of intermediate embedding injection is based on the arithmetic property in the text embedding. Thus, it is important to preserve the token-level information while matching the distribution. 

The histogram in Fig. \ref{supp_fig:dist_analysis}{(Left)} displays the distribution of $c_h$ (blue) and $c_f$ (orange), which shows a distributional mismatch with Wasserstein distance 0.20.
Fig. \ref{supp_fig:dist_analysis}{(Middle)} shows that after final layer normalization, the two distributions become closely aligned, reducing the Wasserstein distance to 0.05. We randomly selected 100 prompts in OAO-AttackBench for the distribution analysis. 
$Mean\text{-}Std(c_h)$ achieves a tighter Wasserstein distance of 0.03. The question is what is lost in the process. 
To better understand this, we plot token-wise cosine similarity between $c_f$ and $LN(c_h)$ (blue) and $Mean\text{-}Std(c_h)$ (purple) in Fig. \ref{supp_fig:dist_analysis}{(Right)}, using the prompt ``Planet Earth is visually defined by purple ocean and glowing continents''. 
$Mean\text{-}Std(c_h)$ shows flatter similarity across the tokens, suggesting that the per-token distinctiveness has been smoothed along with the distribution matching. 
On the other hand, $LN(c_h)$ shows a more fluctuating tendency, suggesting the prompt's word-level information preservation. 
This is essential to introduce the overlooked concept (e.g., attacking attribute) via heavily learned priors (e.g., OAO entities).

\section{OAO-AttackBench}
\label{supp_sec:oao}

\noindent

\begin{wrapfigure}{r}{0.5\textwidth}
\vspace{-2.5em}
\includegraphics[width=\linewidth]{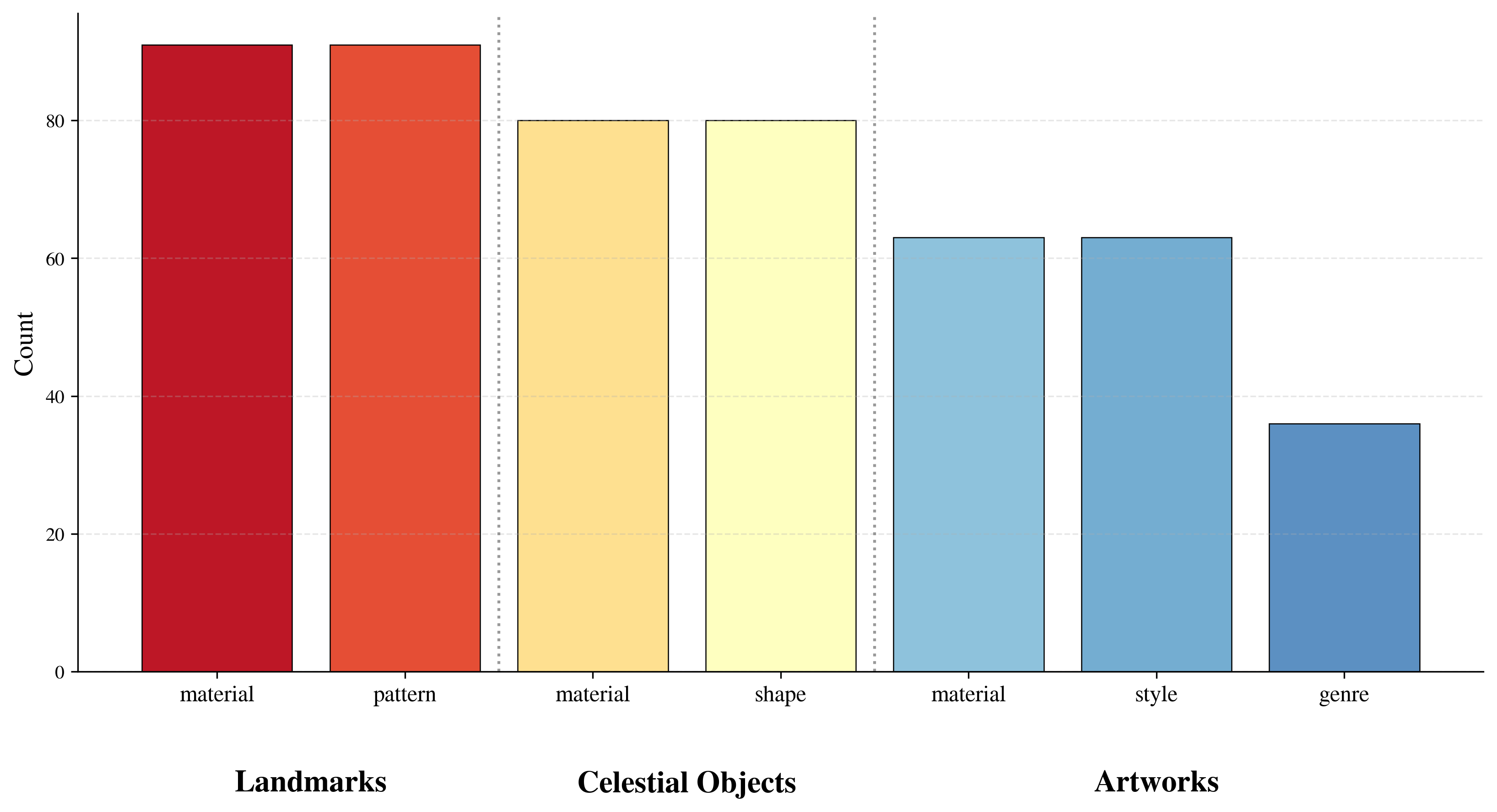}
\caption{The number of samples for each category of OAO-AttackBench.}
\label{supp_fig:oao_stats}
\vspace{-3em}
\end{wrapfigure}

OAO-AttackBench is designed to evaluate a unique setting of creative text prompts for one-and-only (OAO) objects (i.e., entities that exist only in one canonical form). We consider three categories of OAO objects: celestial objects, landmarks, and artworks. 
The full list of OAO objects curated in OAO-AttackBench is provided in Table \ref{supp_tab:oao_objects}. 
We query an LLM to produce attributes that pose an attack for OAO objects. First, we prompt the LLM to list characteristics of the given category. Then, we instruct it to replace those general attributes with unique attributes that are irrelevant, contradict, or violate the core identity. 
The sample LLM instruction is provided in Table \ref{supp_table:llm_prompt}. We use ChatGPT-5.2 to generate attributes.
The resulting attributes and example text prompts included in OAO-AttackBench are shown in Table \ref{supp_tab:oao_attr}. 
The statistics for each category and subcategory are shown in Figure \ref{supp_fig:oao_stats}.

\begin{table}[tb]
  \caption{Full list of one-and-only entities in OAO-AttackBench.
  }
  \label{supp_tab:oao_objects}
  \centering
  \begin{tabular}{@{}l p{0.7\linewidth}@{}}
    \toprule
    Celestial Objects & Sun, Mercury, Venus, Earth, Mars, Jupiter, Saturn, Uranus, Neptune, Moon\\
    \midrule
    Landmarks  & Angkor Wat, The Great Wall, Pyramids of Giza, Great Sphinx of Giza, Sydney Opera House, Big Ben, Statue of Liberty, Taj Mahal, The Colosseum, The Eiffel Tower, Louvre Pyramid, Machu Picchu, Leaning Tower of Pisa\\
    \midrule
    Artworks & Starry Night, Girl with a Pearl Earring, Salvator Mundi, Mona Lisa, Last Supper, Poppies at Argenteuil, Michelangelo's David, Great Wave off Kanagawa, Sunday Afternoon on the Island of La Grande Jatte\\

  \bottomrule
  \end{tabular}
\end{table}

\begin{table*}[h]
\small
\centering
\caption{Sample LLM instruction for OAO-AttackBench.}
\vspace{-0.3cm}
\begin{tabular}{p{0.9\linewidth}}
\toprule
For the given category, 
\\
\\
1. Identify two global characteristics in the following subcategory: shape and material  \\
2. Replace each attribute with a visually or functionally unrelated attribute that contradicts or violates the core identity. \\
Output format: [subcategory], [replaced attribute] \\
\\
The replaced attribute must be applicable to the global characteristics of the given category of objects. 
\\
Category: Celestial objects (e.g., The Sun, The Earth, Mercury, ...)
\\
\bottomrule
\end{tabular}
\vspace{-0.3cm}
\label{supp_table:llm_prompt}
\end{table*}

\begin{table}[tb]
  \caption{Full list of attributes in OAO-AttackBench.}
  \label{supp_tab:oao_attr}
  \centering
  \begin{tabular}{@{}c | >{\centering\arraybackslash}p{0.5\linewidth} | >{\centering\arraybackslash}p{0.35\linewidth}@{}}
  \toprule
  \textbf{Category} & \textbf{Attributes} & \textbf{Examples} \\ 
  \midrule
  \multirow{4}{*}{Material}
  & Aluminium, Concrete, Silver, Liquid, Cotton, Plastic, Wood, Glass, Jelly, Lava, Clay, Optic fiber, Wires, Snow, Fabric, Knit, Cardboard, Leather, Crystal
& \multirow{4}{=}{\centering Machu Picchu made of clay, \\ Cotton Earth, \\ Knitted Mona Lisa} \\
\midrule
  \multirow{2}{*}{Shape}
  & Heart, Hexagon, Pentagon, Octagon, Cylinder, Cube, Donut, Square
  & \multirow{2}{*}{Octagon-shaped Earth} \\
  \midrule
  \multirow{2}{*}{Pattern}
  & Flower, Heart, Paisley, Stripe, Checkered, Polka-dotted, Cloud
  & \multirow{2}{*}{Flower-patterned Big Ben} \\
  \midrule
  \multirow{2}{*}{Genre}
  & \multirow{2}{*}{Stage Show, Movie, Mural, Cartoon}
  & Stage Show version of Starry Night \\
  \midrule
  \multirow{2}{*}{Style}
  & Dragon, Boat, New York Skyline, Dog, Cat, Airplane, Person
  & \multirow{2}{*}{Starry-Night-style person} \\
  \bottomrule
  \end{tabular}
\end{table}

\section{Implementation Details}
\label{supp_sec:implementation_details}

Stable Diffusion (SD) 3.0~\cite{esser2024scaling} has three text encoders (OpenCLIP ViT-G, CLIP ViT-L, and T5-XXL) and two types of text embeddings (token embeddings and pooled embeddings). The outputs of the three text encoders form the final token and pooled embeddings used by the diffusion model.
In our approach, we extract intermediate representations from all three text encoders and use them to construct the token embedding. For the pooled embedding, we retain the final representation produced by each encoder. The resulting token and pooled embeddings are then concatenated and fed to the denoising network.
For both SD2.1 and SD3.0, we set the injection parameter $\lambda$ to 0.2.

\section{Evolution of Intermediate Representations}
\label{supp_sec:qualitativeepth_study}

\begin{wrapfigure}{r}{0.5\textwidth}
\vspace{-1em}
\includegraphics[width=\linewidth]{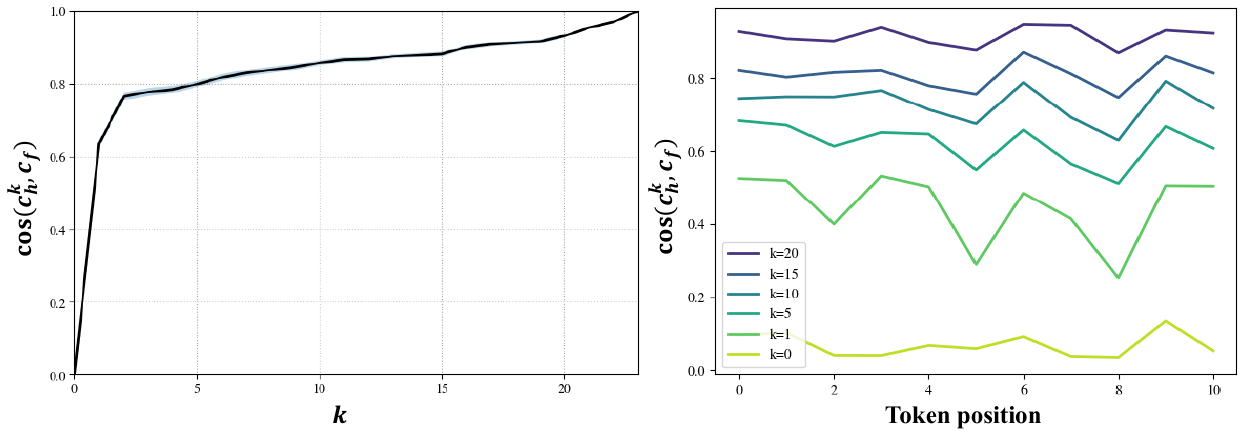}
\caption{Layer-wise \textbf{(left)} and token-wise \textbf{(right)} cosine similarity to the final embedding across depth. Both are non-linear: deeper layers converge toward sentence-level semantics; earlier layer representations retain token-level distinctiveness.}
\label{supp_fig:depth_study}
\vspace{-1em}
\end{wrapfigure}

In this section, we present the analysis of the evolution of the intermediate representations of the text encoder. 

Fig. \ref{supp_fig:depth_study} displays the layer-wise (left) and token-wise (right) cosine similarity between the final representation ($c_f$) and the intermediate representation at layer $k$ ($c_h^k$) in SD2.1's text encoder with the prompt ``Planet Earth is visually defined by purple ocean and glowing continents''.

Fig. \ref{supp_fig:depth_study} (left) shows that the similarity saturates rapidly within the first few layers and plateaus afterwards, indicating that the evolution of intermediate text representations is \textit{far from linear}. In other words, most of the semantic transformation occurs in a narrow band of early-to-mid layers, after which representations remain largely stable. 

In Fig. \ref{supp_fig:depth_study} (right), 
we observe that token-wise cosine similarity across transformer depth is also non-linear, and the per-token differences are progressively smoothed as depth increases. Taken together, these two observations support a consistent picture of the encoder's behavior that deeper layers converge toward global, sentence-level semantics, while earlier layers retain token-level distinctiveness. 
OAO alignments need a word-level signal that early layers preserve, hence providing deeper understanding of the ablation study on the layer choice in \S 4.3.

\section{Qualitative Results}
\label{supp_sec:qualitative}

\begin{figure}[t]
  \centering
  \includegraphics[width=0.9\textwidth]{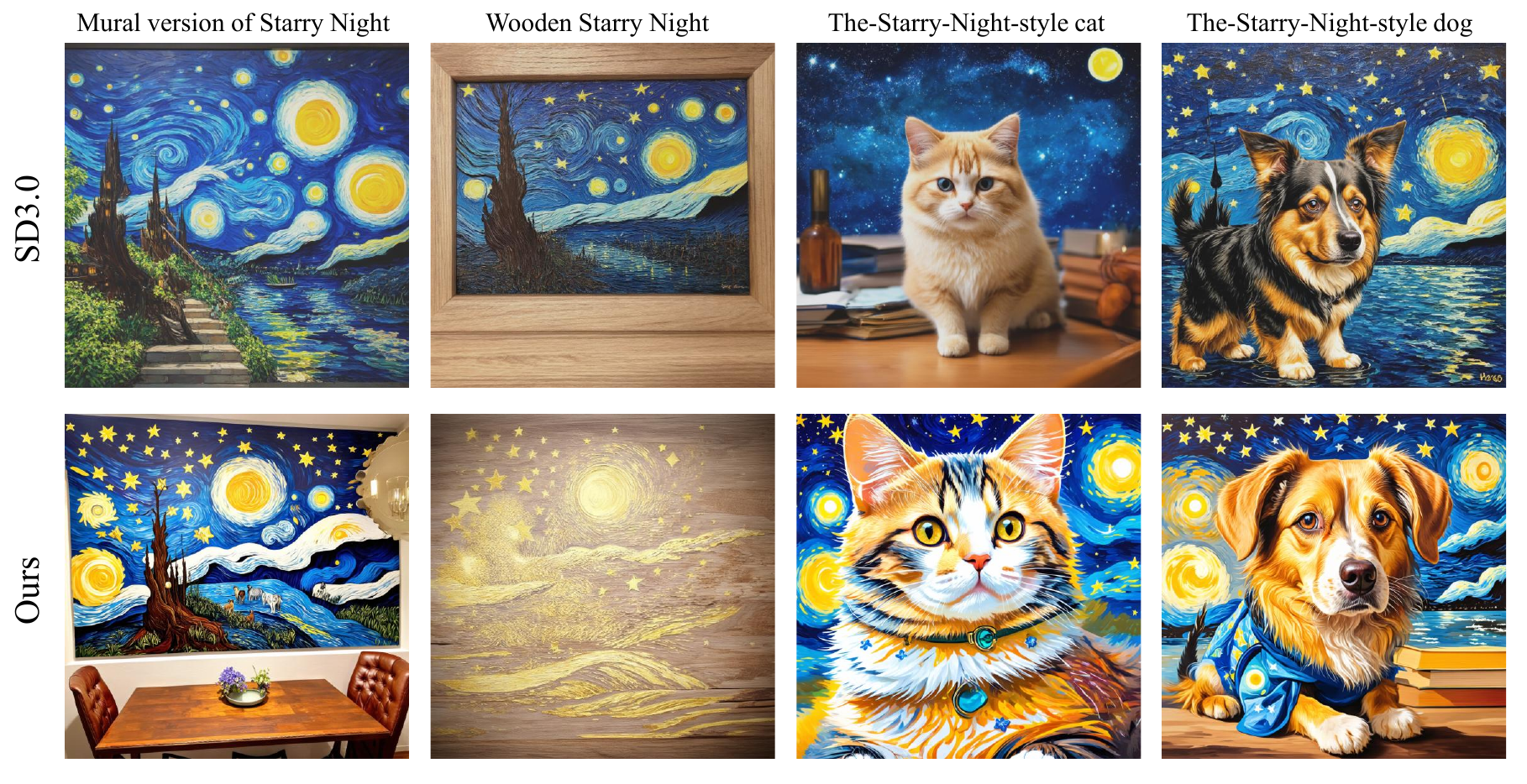}
  \caption{
  Samples generated from SD3.0 and IR-guidance on text prompts that include ``The Starry Night''.
  }
  \label{supp_fig:quali-starry}
\end{figure}

\begin{figure}[t]
  \centering
  \includegraphics[width=0.9\textwidth]{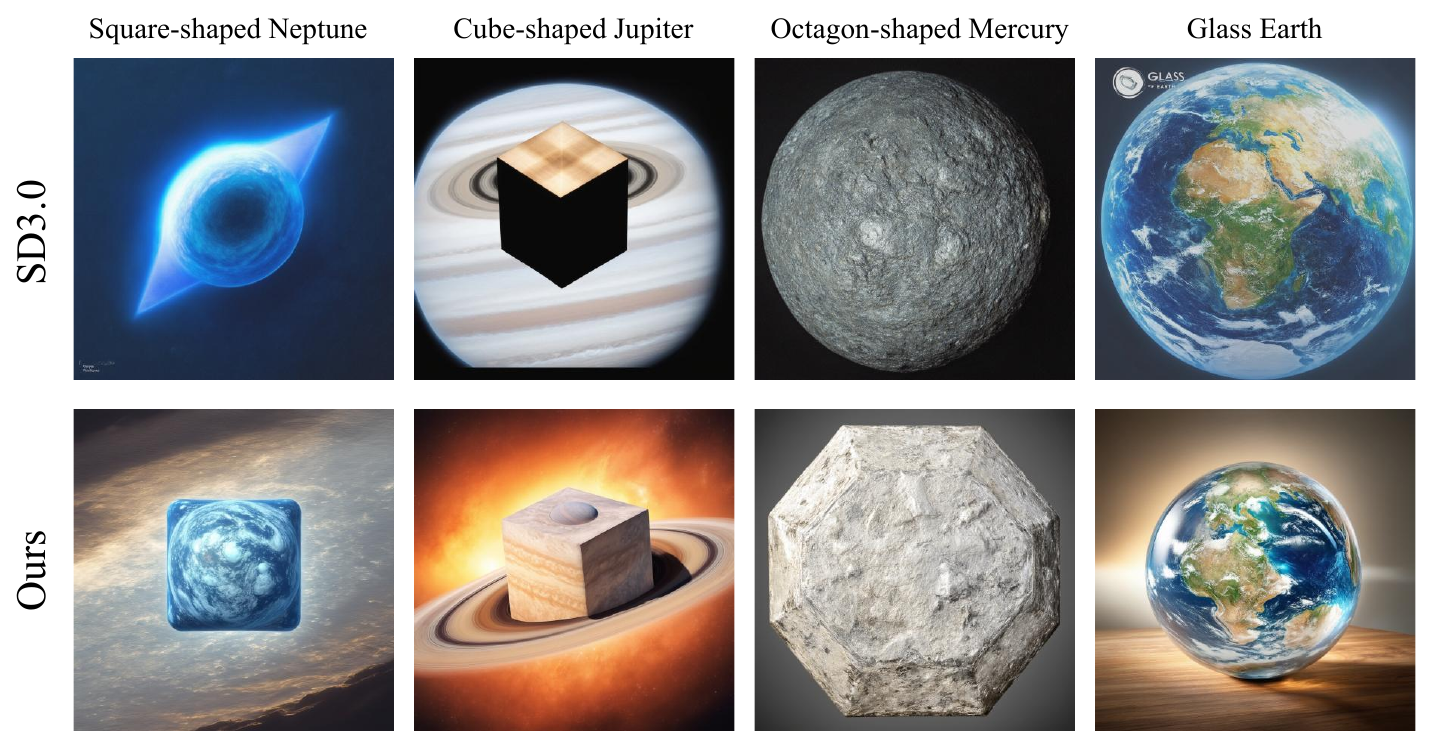}
  \caption{Samples generated from SD3.0 and IR-guidance on text prompts that include celestial objects. 
  }
  \label{supp_fig:quali-celestial}
\end{figure}

\begin{figure}[t]
  \centering
  \includegraphics[width=0.9\textwidth]{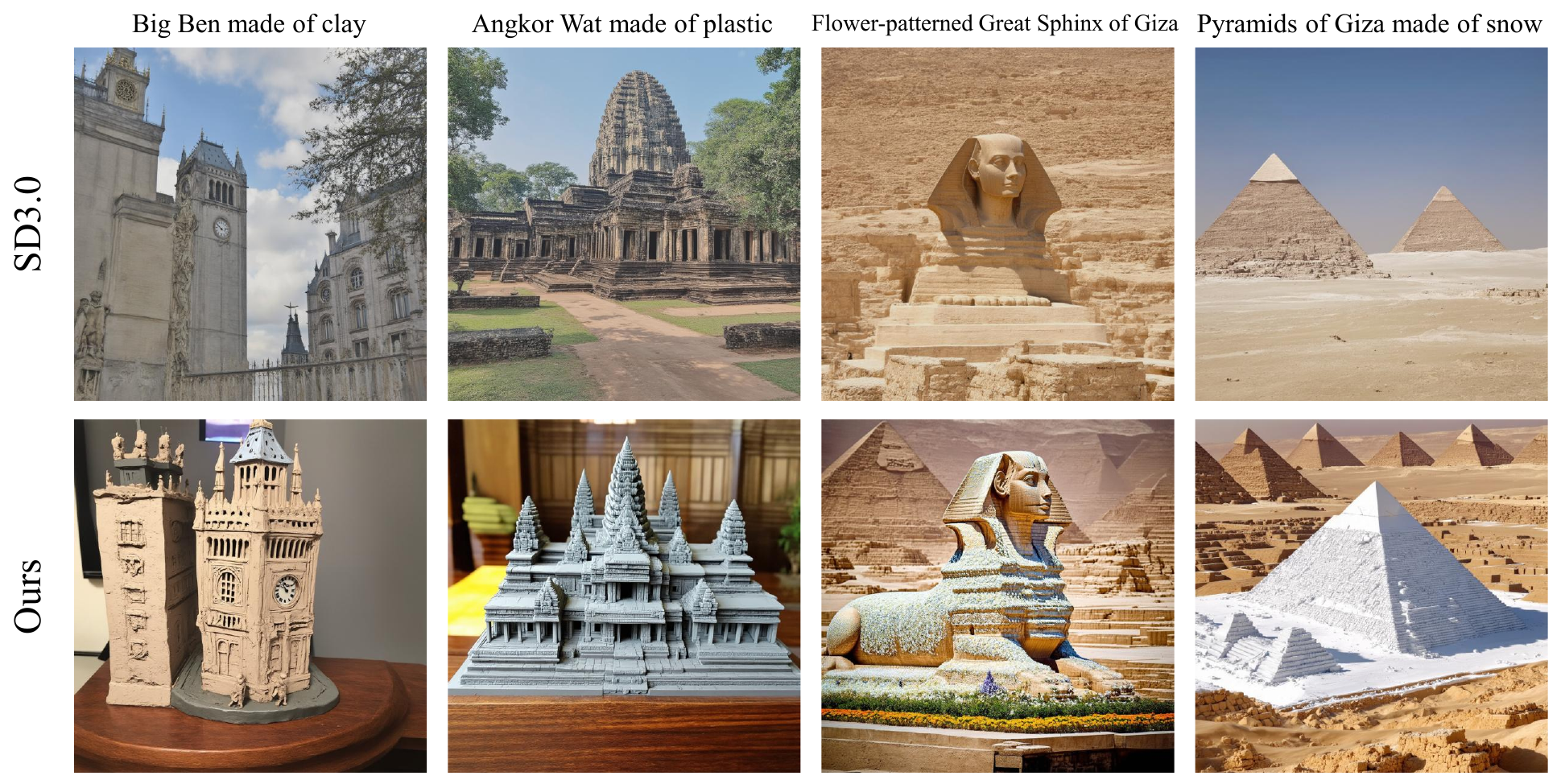}
  \caption{Samples generated from SD3.0 and IR-guidance on text prompts that include landmarks.
  }
  \label{supp_fig:quali-landmarks}
\end{figure}

\noindent
\textbf{More Qualitative Results}
We provide attribute variants of \textit{Starry Night} in Figure \ref{supp_fig:quali-starry}. Baseline SD3.0 fails to associate the explicit concept given in the text with The Starry Night. Generated images from IR-guidance depict the given text prompt more faithfully. We provide more qualitative samples from celestial objects and landmarks in Figure \ref{supp_fig:quali-celestial} and Figure \ref{supp_fig:quali-landmarks}, respectively.

\section{Ablation Study}
\label{supp_sec:proofbl}

\begin{figure}[h]
\centering

\begin{minipage}[t]{0.41\linewidth}
\centering
\includegraphics[width=\linewidth]{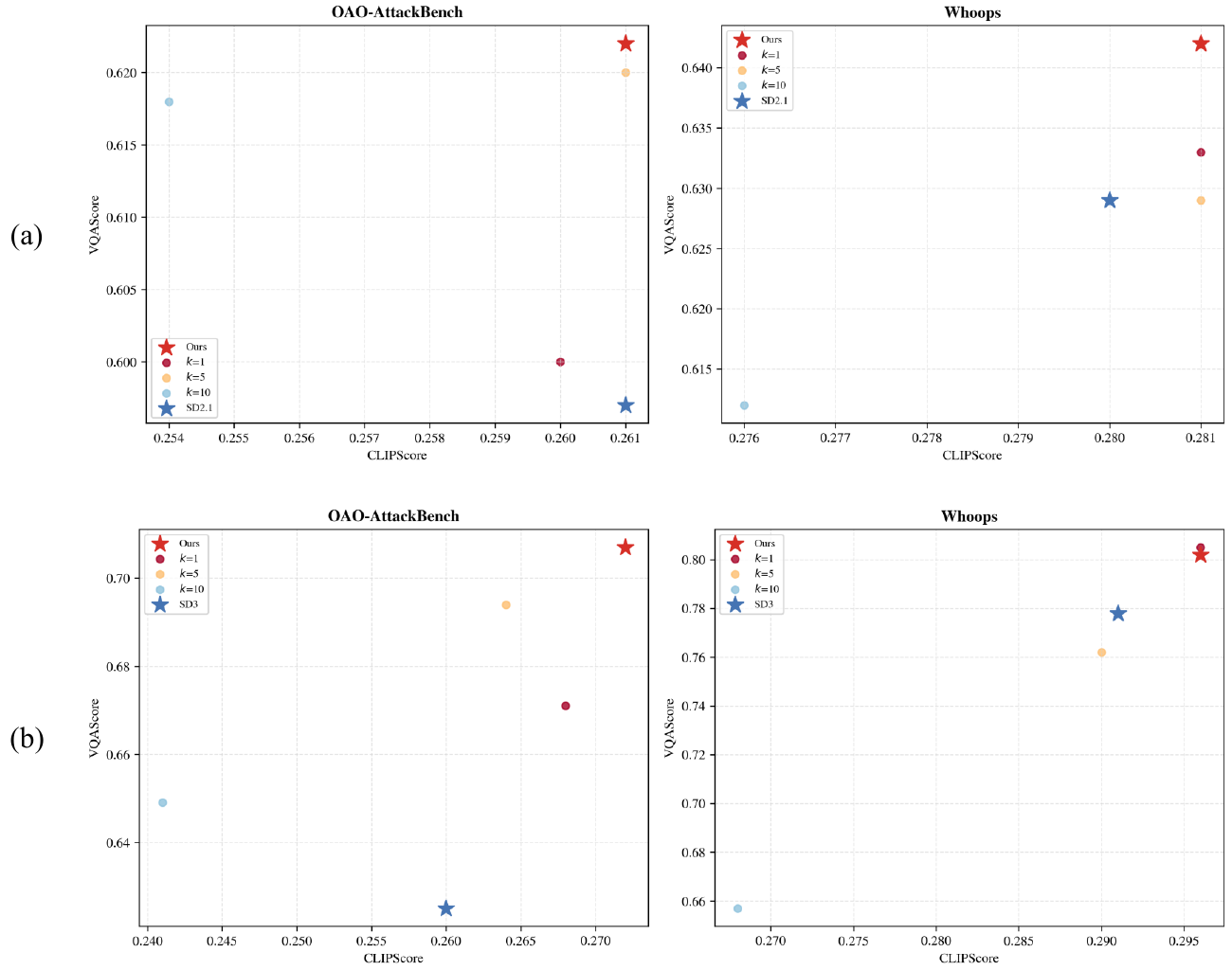}
\caption{Performance under fixed scheduling and IR-guided diffusion (ours) on OAO-AttackBench and Whoops for (a) SD2.1 and (b) SD3.0.}
\label{supp_fig:repX}
\end{minipage}
\hspace{0.03\linewidth}
\begin{minipage}[t]{0.41\linewidth}
\centering
\includegraphics[width=\linewidth]{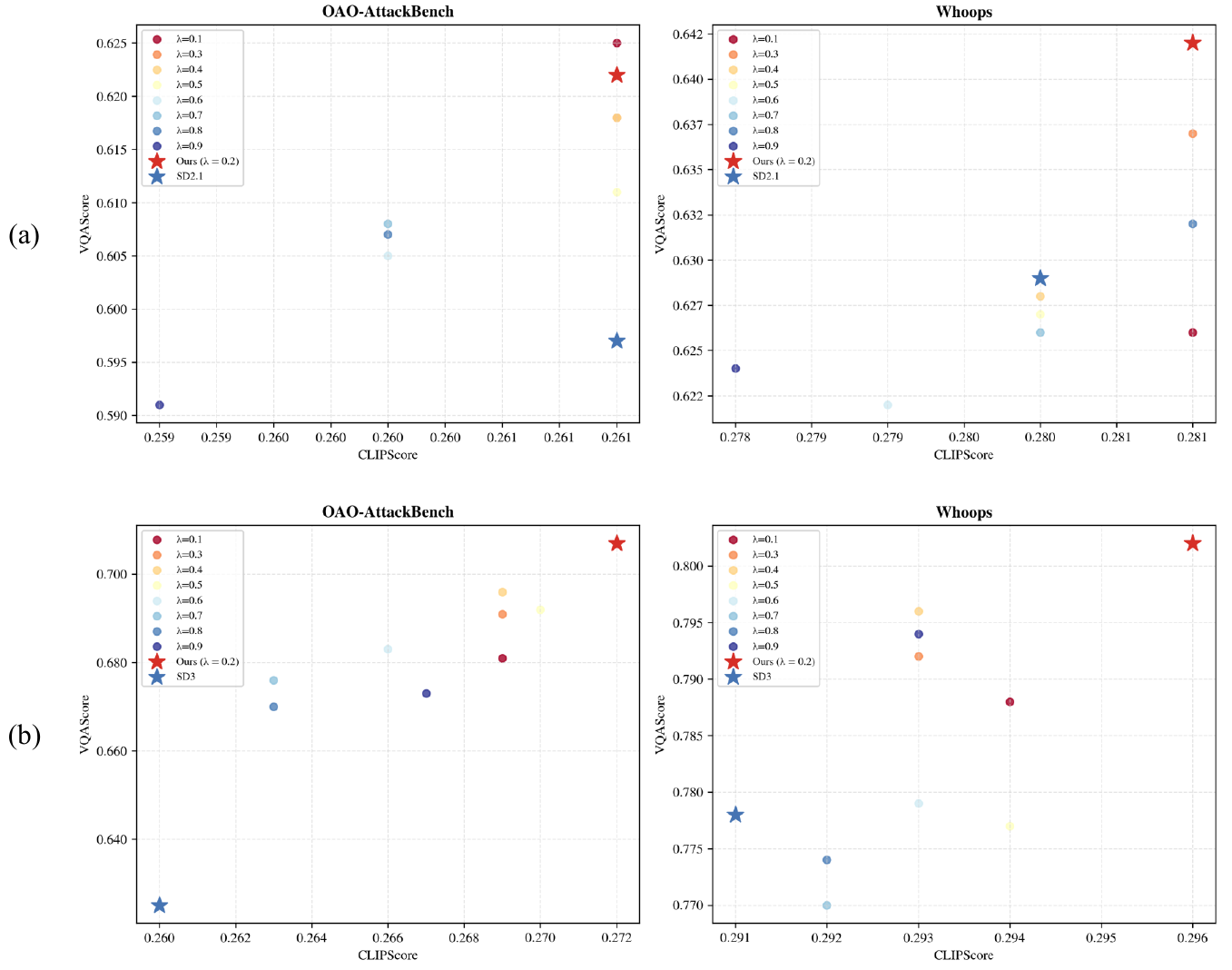}
\caption{Performance under various injection strengths on OAO-AttackBench and Whoops for (a) SD2.1 and (b) SD3.0.}
\label{supp_fig:lamdX}
\end{minipage}

\vspace{-0.6em}
\end{figure}

\subsection{Fixed Scheduling} 
In this section, we present ablation studies on the effectiveness of adaptive scheduling in IR-guided diffusion described in Section 3.4. 
In Figure \ref{supp_fig:repX}, we observe that IR-guidance with adaptive scheduling (denoted by a red star \textcolor{red}{$\bigstar$}) achieves the best overall performance. 
We highlight that IR-guided diffusion consistently outperforms the fixed-schedule variants, while under fixed scheduling, the optimal number of guidance steps varies across datasets and architectures.
For example, SD2.1 on OAO-AttackBench (Fig. \ref{supp_fig:repX}(a)), among the fixed-schedule variants, $k=5$ achieves the best performance, while $k=10$ gives the second-best VQAScore, and $k=1$ yields the second-best CLIPScore, where $k$ denotes the timestep IR-guidance is applied.
Whoops tells a different story. Here, $k=1$ achieves  the optimal performance among fixed scheduling variants. While $k=1$ slightly outperforms IR-guided diffusion in terms of VQAScore, we emphasize that IR-guided diffusion shows consistently strong performance across datasets and architectures, whereas the effectiveness of fixed scheduling varies.
In other words, the optimal fixed scheduling step depends on the specific model architecture and dataset. In contrast, our method provides a more stable strategy by adaptively determining the scheduling, removing the need for manual parameter tuning for each architecture or dataset.

\subsection{Injection Strength}
\label{supp_sec:injection_strength}
In Section 3.3, we define IR-guided embedding as
\begin{equation}
    \label{supp_eq:injection}
    \tilde{c} = c_f + \lambda c_h ,
\end{equation}
where $c_f$ is the final text embedding, $c_h$ is the representation at an intermediate block, and $\lambda$ is the injection strength. 
In this section, we present the ablation results on the injection strength in Figure \ref{supp_fig:lamdX}. 
We find that a small injection strength $\lambda \in [0.1, 0.5]$ generally yields higher performance.

\section{Limitations and Future Work}
\label{supp_sec:F}

In the first column of Figure \ref{supp_fig:quali-landmarks}, the image generated for ``Big Ben made of clay'' with IR-guidance exhibits a background that better aligns with the notion of ``made of clay'', suggesting that IR-guidance introduces visual features associated with the attribute ``clay''.
A similar phenomenon is observed on the second column of Figure \ref{supp_fig:quali-landmarks}. 
As discussed in \S 4.2 (see Figure 4(b)), adjusting the overall context to the specific features of the text prompt can result in more faithful images; however, it may result in unwanted context.
Note that applying modifications only to the target object falls under a different research problem: Text-to-image editing \cite{wu2025nep, sajnani2025geodiffuser, hertz2022prompt}. The goal of IR-guidance is to generate images faithful to the text prompt in settings where no reference image exists.
Nevertheless, exploring image editing for OAO objects is a promising direction for future work.

\begin{wrapfigure}{r}{0.45\textwidth}
\vspace{-2em}
\includegraphics[width=\linewidth]{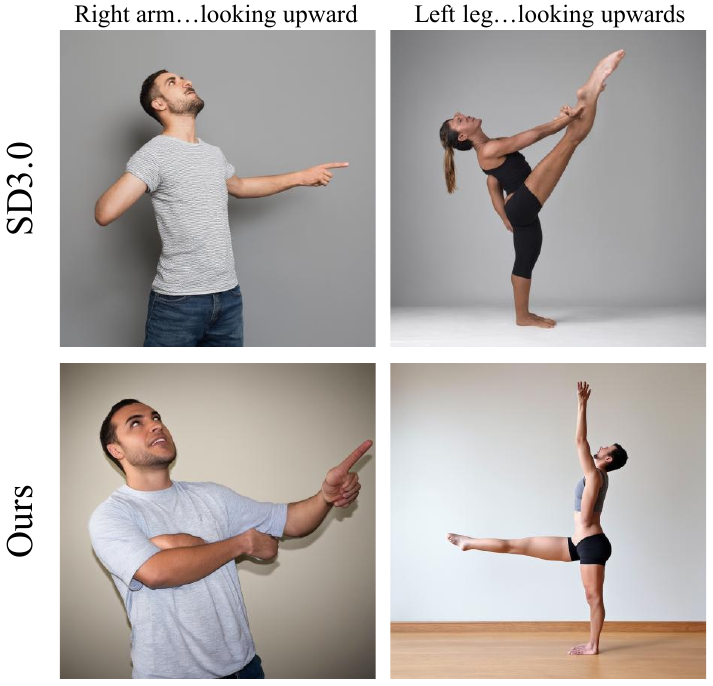}
\caption{IR-guidance inherits the limitations of the baseline SD3.0 on complex human anatomy.}
\label{supp_fig:human_pose}
\vspace{-5mm}
\end{wrapfigure}

Additionally, IR-guidance inherits the backbone's limitations on human anatomy. 
Fig. \ref{supp_fig:human_pose} displays generated images from the prompt ``Right arm pointing behind almost straight back left arm straight in front and crosses a little over chest face looking upward'' (left), and ``Left leg is lifted behind body with knee bent and left arm up extended out at face level. The right arm is lifted up to chin level and the head is looking upwards'' (right). It shows that IR-guidance cannot fully address the limitations of the backbone to generate anatomically plausible complex human poses.

\section{Additional Metrics}
\label{supp_sec:proofdditional_metrics}

\begin{table}[h]
\centering
\caption{Quantitative comparison on OAO-AttackBench and Whoops. Bold values denote the best score.}
\label{supp_tab:additional_metrics}

\vspace{-10pt}
\resizebox{\linewidth}{!}{
\begin{tabular}{llccccc|ccccc}
\toprule
& & \multicolumn{5}{c|}{OAO-AttackBench} 
& \multicolumn{5}{c}{Whoops} \\
\cmidrule(lr){3-7} \cmidrule(lr){8-12}
 & 
& {\fontsize{7}{8}\selectfont CLIP$\uparrow$}
& {\fontsize{7}{8}\selectfont VQA$\uparrow$}
& {\fontsize{7}{8}\selectfont VE HR$\uparrow$}
& {\fontsize{7}{8}\selectfont V C-HR$\uparrow$}
& {\fontsize{7}{8}\selectfont VE C-HR$\uparrow$}
& {\fontsize{7}{8}\selectfont CLIP$\uparrow$}
& {\fontsize{7}{8}\selectfont VQA$\uparrow$}
& {\fontsize{7}{8}\selectfont CHAIR-I$\downarrow$}
& {\fontsize{7}{8}\selectfont CHAIR-S$\downarrow$}
& {\fontsize{7}{8}\selectfont Recall$\uparrow$} \\
\midrule

\multirow{2}{*}{SD2.1}
& baseline
& \textbf{0.261} & 0.596 & 0.667 & 0.631 & 0.546
& 0.280 & 0.629 & \textbf{0.664} & 0.978 & \textbf{0.397} \\

& ours
& \textbf{0.261} & \textbf{0.621} & \textbf{0.705} & \textbf{0.642} & \textbf{0.571}
& \textbf{0.281} & \textbf{0.641} & 0.681 & \textbf{0.972} & 0.384 \\

\midrule

\multirow{2}{*}{SD3.0}
& baseline
& 0.260 & 0.625 & 0.695 & 0.689 & 0.610
& 0.291 & 0.778 & 0.616 & 0.960 & 0.456 \\

& ours
& \textbf{0.272} & \textbf{0.707} & \textbf{0.798} & \textbf{0.744} & \textbf{0.691}
& \textbf{0.296} & \textbf{0.802} & \textbf{0.611} & \textbf{0.950} & \textbf{0.476} \\

\bottomrule
\end{tabular}
}
\vspace{-12pt}
\end{table}

Table \ref{supp_tab:additional_metrics} reports additional quantitative results on OAO-AttackBench and Whoops. 
For additional metrics, we adopt graph structure decomposition of texts to dissect the text into central concepts and their dependencies \cite{cho2024davidsonian}. Then, we query the VQA model to evaluate if the element exists in the generated image. It enables fine-grained assessment of individual concepts as well as their compositionality. 
Experimental results show that IR-guidance consistently outperforms baseline under additional metrics. 

On OAO-AttackBench, VE HR denotes Vertex-Edge Hit Rate where we average the hit rate of each element. It is considered hit (1) if the element exists in the image (the output score of VQA model is above 0.5), and miss (0) otherwise. 
V C-HR (Vertex Cascading-HitRate) is where only the vertex is considered, but in a cascading way. Given the nature of OAO-AttackBench, the prompt can be divided into the central object (OAO) and the attacking attribute. Both OAO and the attribute are considered vertices, but the attribute is dependent on OAO. Using this, V C-HR examines OAO hit rate first, then examines the attribute hit only if the OAO was a hit. VE C-HR (Vertex-Edge Cascading-Hit-Rate) is the most challenging metric where we examine edge hit only if both vertices that correspond to the edge are hits. 

On Whoops, it is difficult to dissect the prompt in a graph form, as we do not have prior knowledge of what the central object is and what the attribute is, and how they are connected. Thus, we employ grounded captioning-based metrics (CHAIR \cite{rohrbach2018object} variants) and recall. Experimental results consistently demonstrate the effectiveness of our method. 

\section{Failure Cases on Closed-source Models}
\label{supp_sec:closed-sourced}

\begin{wrapfigure}{r}{0.45\textwidth}
\vspace{-1.5em}
\includegraphics[width=\linewidth]{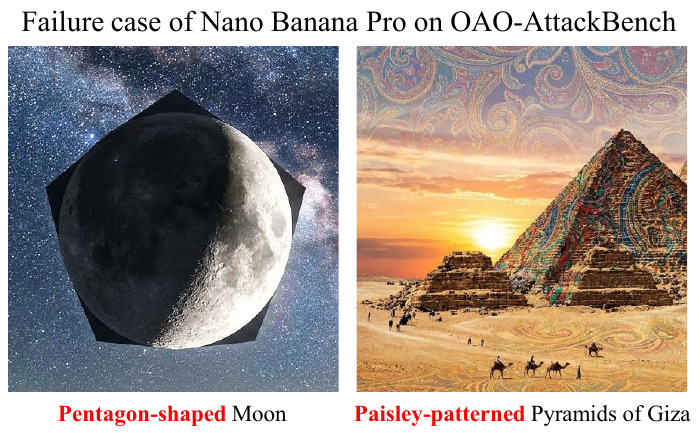}
\caption{Nano Banana Pro  (gemini-3-pro-image-preview) fails on OAO-AttackBench}
\label{supp_fig:nanobanana}
\vspace{-2em}
\end{wrapfigure}

We employed Nano Banana Pro on OAO-AttackBench. The VQAScore is 0.819, scoring highest on landmarks (0.84), and lowest on celestial objects (0.79).
However, Nano Banana Pro still suffers from shape-attacked prompts on celestial objects. Fig. \ref{supp_fig:nanobanana} (left) shows an arbitrary pentagon-shaped figure, rather than a pentagonal Moon.
Also, Fig. \ref{supp_fig:nanobanana} (right) displays paisley pattern throughout the image, rather than only on Pyramids, suggesting object-attribute binding errors still persist in OAO entities.
Overall, Gemini results show that 
concept association bias persists in frontier closed-source models, confirming that OAO alignment remains an open problem.


\begin{wraptable}{r}{0.42\textwidth}
\vspace{-3.5em}
\centering
\caption{Computational cost comparison.}
\label{tab:computational_cost}
\small
\begin{tabular}{lcc}
\hline
 & \makecell{GPU Time\\(sec)} & \makecell{Peak Mem.\\(GB)} \\
\hline
SD3.0 & 4.80 & 18.76 \\
Ours & 5.07 & 18.76 \\
\hline
\end{tabular}
\vspace{-15pt}
\end{wraptable}

\section{Runtime \& Memory}
\label{supp_sec:runtime_n_memory}
To demonstrate the efficiency of our method, 
we compare GPU time (sec) and peak memory (GB) of SD3.0 and our method on an A100 GPU averaged over 10 random prompts of OAO-AttackBench, including the text embedding manipulation part. The results are summarized in Table \ref{tab:computational_cost}.

%% file: main.bib
@String(TOG   = {ACM Trans. Graph.})

@String(TOG   = {ACM TOG})

@inproceedings{lemon,
  title={When are lemons purple? the concept association bias of vision-language models},
  author={Tang, Yingtian and Yamada, Yutaro and Zhang, Yoyo and Yildirim, Ilker},
  booktitle={Proceedings of the 2023 Conference on Empirical Methods in Natural Language Processing},
  pages={14333--14348},
  year={2023}
}

@article{toker2024diffusion,
  title={Diffusion lens: Interpreting text encoders in text-to-image pipelines},
  author={Toker, Michael and Orgad, Hadas and Ventura, Mor and Arad, Dana and Belinkov, Yonatan},
  journal={arXiv preprint arXiv:2403.05846},
  year={2024}
}

@inproceedings{kim2025text,
  title={Text embedding is not all you need: Attention control for text-to-image semantic alignment with text self-attention maps},
  author={Kim, Jeeyung and Esmaeili, Erfan and Qiu, Qiang},
  booktitle={Proceedings of the Computer Vision and Pattern Recognition Conference},
  pages={8031--8040},
  year={2025}
}

@article{li2025does,
  title={Does object binding naturally emerge in large pretrained vision transformers?},
  author={Li, Yihao and Salehi, Saeed and Ungar, Lyle and Kording, Konrad P},
  journal={arXiv preprint arXiv:2510.24709},
  year={2025}
}

@article{gecko,
  title={Revisiting text-to-image evaluation with gecko: On metrics, prompts, and human ratings},
  author={Wiles, Olivia and Zhang, Chuhan and Albuquerque, Isabela and Kaji{\'c}, Ivana and Wang, Su and Bugliarello, Emanuele and Onoe, Yasumasa and Papalampidi, Pinelopi and Ktena, Ira and Knutsen, Chris and others},
  journal={arXiv preprint arXiv:2404.16820},
  year={2024}
}

@inproceedings{tifa,
  title={Tifa: Accurate and interpretable text-to-image faithfulness evaluation with question answering},
  author={Hu, Yushi and Liu, Benlin and Kasai, Jungo and Wang, Yizhong and Ostendorf, Mari and Krishna, Ranjay and Smith, Noah A},
  booktitle={Proceedings of the IEEE/CVF International Conference on Computer Vision},
  pages={20406--20417},
  year={2023}
}

@inproceedings{stanford,
  title={A hierarchical approach for generating descriptive image paragraphs},
  author={Krause, Jonathan and Johnson, Justin and Krishna, Ranjay and Fei-Fei, Li},
  booktitle={Proceedings of the IEEE conference on computer vision and pattern recognition},
  pages={317--325},
  year={2017}
}

@inproceedings{localized_narrative,
  title={Connecting vision and language with localized narratives},
  author={Pont-Tuset, Jordi and Uijlings, Jasper and Changpinyo, Soravit and Soricut, Radu and Ferrari, Vittorio},
  booktitle={European conference on computer vision},
  pages={647--664},
  year={2020},
  organization={Springer}
}

@inproceedings{countbench,
  title={Teaching clip to count to ten},
  author={Paiss, Roni and Ephrat, Ariel and Tov, Omer and Zada, Shiran and Mosseri, Inbar and Irani, Michal and Dekel, Tali},
  booktitle={Proceedings of the IEEE/CVF international conference on computer vision},
  pages={3170--3180},
  year={2023}
}

@inproceedings{vrd,
  title={Visual relationship detection with language priors},
  author={Lu, Cewu and Krishna, Ranjay and Bernstein, Michael and Fei-Fei, Li},
  booktitle={European conference on computer vision},
  pages={852--869},
  year={2016},
  organization={Springer}
}

@inproceedings{diffusiondb,
  title={Diffusiondb: A large-scale prompt gallery dataset for text-to-image generative models},
  author={Wang, Zijie J and Montoya, Evan and Munechika, David and Yang, Haoyang and Hoover, Benjamin and Chau, Duen Horng},
  booktitle={Proceedings of the 61st annual meeting of the association for computational linguistics (volume 1: Long papers)},
  pages={893--911},
  year={2023}
}

@article{posescript,
  title={Posescript: Linking 3d human poses and natural language},
  author={Delmas, Ginger and Weinzaepfel, Philippe and Lucas, Thomas and Moreno-Noguer, Francesc and Rogez, Gr{\'e}gory},
  journal={IEEE transactions on pattern analysis and machine intelligence},
  year={2024},
  publisher={IEEE}
}

@inproceedings{drawtext,
  title={Character-aware models improve visual text rendering},
  author={Liu, Rosanne and Garrette, Dan and Saharia, Chitwan and Chan, William and Roberts, Adam and Narang, Sharan and Blok, Irina and Mical, RJ and Norouzi, Mohammad and Constant, Noah},
  booktitle={Proceedings of the 61st Annual Meeting of the Association for Computational Linguistics (Volume 1: Long Papers)},
  pages={16270--16297},
  year={2023}
}

@inproceedings{whoops,
  title={Breaking common sense: Whoops! a vision-and-language benchmark of synthetic and compositional images},
  author={Bitton-Guetta, Nitzan and Bitton, Yonatan and Hessel, Jack and Schmidt, Ludwig and Elovici, Yuval and Stanovsky, Gabriel and Schwartz, Roy},
  booktitle={Proceedings of the IEEE/CVF International Conference on Computer Vision},
  pages={2616--2627},
  year={2023}
}

@misc{mj,
  title={Midjourney user prompts \& generated images (250k)},
  author={Turc, Iulia and Nemade, Gaurav},
  year={2023}
}

@inproceedings{vqascore,
  title={Evaluating text-to-visual generation with image-to-text generation},
  author={Lin, Zhiqiu and Pathak, Deepak and Li, Baiqi and Li, Jiayao and Xia, Xide and Neubig, Graham and Zhang, Pengchuan and Ramanan, Deva},
  booktitle={European Conference on Computer Vision},
  pages={366--384},
  year={2024},
  organization={Springer}
}

@inproceedings{clipscore,
  title={Clipscore: A reference-free evaluation metric for image captioning},
  author={Hessel, Jack and Holtzman, Ari and Forbes, Maxwell and Le Bras, Ronan and Choi, Yejin},
  booktitle={Proceedings of the 2021 conference on empirical methods in natural language processing},
  pages={7514--7528},
  year={2021}
}

@article{hps,
  title={Human preference score v2: A solid benchmark for evaluating human preferences of text-to-image synthesis},
  author={Wu, Xiaoshi and Hao, Yiming and Sun, Keqiang and Chen, Yixiong and Zhu, Feng and Zhao, Rui and Li, Hongsheng},
  journal={arXiv preprint arXiv:2306.09341},
  year={2023}
}

@article{wang2023diffusion,
  title={Diffusion models generate images like painters: an analytical theory of outline first, details later},
  author={Wang, Binxu and Vastola, John J},
  journal={arXiv preprint arXiv:2303.02490},
  year={2023}
}

@article{chefer2023attend,
  title={Attend-and-excite: Attention-based semantic guidance for text-to-image diffusion models},
  author={Chefer, Hila and Alaluf, Yuval and Vinker, Yael and Wolf, Lior and Cohen-Or, Daniel},
  journal={ACM transactions on Graphics (TOG)},
  volume={42},
  number={4},
  pages={1--10},
  year={2023},
  publisher={ACM New York, NY, USA}
}

@article{hertz2022prompt,
  title={Prompt-to-prompt image editing with cross attention control},
  author={Hertz, Amir and Mokady, Ron and Tenenbaum, Jay and Aberman, Kfir and Pritch, Yael and Cohen-Or, Daniel},
  journal={arXiv preprint arXiv:2208.01626},
  year={2022}
}

@article{efron2011tweedie,
  title={Tweedie’s formula and selection bias},
  author={Efron, Bradley},
  journal={Journal of the American Statistical Association},
  volume={106},
  number={496},
  pages={1602--1614},
  year={2011},
  publisher={Taylor \& Francis}
}

@incollection{robbins1992empirical,
  title={An empirical Bayes approach to statistics},
  author={Robbins, Herbert E},
  booktitle={Breakthroughs in Statistics: Foundations and basic theory},
  pages={388--394},
  year={1992},
  publisher={Springer}
}

@article{hu2024token,
  title={Token merging for training-free semantic binding in text-to-image synthesis},
  author={Hu, Taihang and Li, Linxuan and Van de Weijer, Joost and Gao, Hongcheng and Shahbaz Khan, Fahad and Yang, Jian and Cheng, Ming-Ming and Wang, Kai and Wang, Yaxing},
  journal={Advances in Neural Information Processing Systems},
  volume={37},
  pages={137646--137672},
  year={2024}
}

@inproceedings{kim2025rethinking,
  title={Rethinking training for de-biasing text-to-image generation: Unlocking the potential of stable diffusion},
  author={Kim, Eunji and Kim, Siwon and Park, Minjun and Entezari, Rahim and Yoon, Sungroh},
  booktitle={Proceedings of the Computer Vision and Pattern Recognition Conference},
  pages={13361--13370},
  year={2025}
}

@misc{li2025fair,
      title={Fair Text-to-Image Diffusion via Fair Mapping}, 
      author={Jia Li and Lijie Hu and Jingfeng Zhang and Tianhang Zheng and Hua Zhang and Di Wang},
      year={2024},
      eprint={2311.17695},
      archivePrefix={arXiv},
      primaryClass={cs.CV},
      url={https://arxiv.org/abs/2311.17695}, 
}

@inproceedings{zhang2025joint,
  title={Joint vision-language social bias removal for clip},
  author={Zhang, Haoyu and Guo, Yangyang and Kankanhalli, Mohan},
  booktitle={Proceedings of the Computer Vision and Pattern Recognition Conference},
  pages={4246--4255},
  year={2025}
}

@article{chuang2023debiasing,
  title={Debiasing vision-language models via biased prompts},
  author={Chuang, Ching-Yao and Jampani, Varun and Li, Yuanzhen and Torralba, Antonio and Jegelka, Stefanie},
  journal={arXiv preprint arXiv:2302.00070},
  year={2023}
}

@article{shen2023finetuning,
  title={Finetuning text-to-image diffusion models for fairness},
  author={Shen, Xudong and Du, Chao and Pang, Tianyu and Lin, Min and Wong, Yongkang and Kankanhalli, Mohan},
  journal={arXiv preprint arXiv:2311.07604},
  year={2023}
}

@article{friedrich2023fair,
  title={Fair diffusion: Instructing text-to-image generation models on fairness},
  author={Friedrich, Felix and Brack, Manuel and Struppek, Lukas and Hintersdorf, Dominik and Schramowski, Patrick and Luccioni, Sasha and Kersting, Kristian},
  journal={arXiv preprint arXiv:2302.10893},
  year={2023}
}

@inproceedings{um2025minority,
  title={Minority-focused text-to-image generation via prompt optimization},
  author={Um, Soobin and Ye, Jong Chul},
  booktitle={Proceedings of the Computer Vision and Pattern Recognition Conference},
  pages={20926--20936},
  year={2025}
}

@inproceedings{rombach2022high,
  title={High-resolution image synthesis with latent diffusion models},
  author={Rombach, Robin and Blattmann, Andreas and Lorenz, Dominik and Esser, Patrick and Ommer, Bj{\"o}rn},
  booktitle={Proceedings of the IEEE/CVF Conference on Computer Vision and Pattern Recognition},
  pages={10684--10695},
  year={2022}
}

@inproceedings{esser2024scaling,
  title={Scaling rectified flow transformers for high-resolution image synthesis},
  author={Esser, Patrick and Kulal, Sumith and Blattmann, Andreas and Entezari, Rahim and M{\"u}ller, Jonas and Saini, Harry and Levi, Yam and Lorenz, Dominik and Sauer, Axel and Boesel, Frederic and others},
  booktitle={Forty-first international conference on machine learning},
  year={2024}
}

@inproceedings{orgad2023editing,
  title={Editing implicit assumptions in text-to-image diffusion models},
  author={Orgad, Hadas and Kawar, Bahjat and Belinkov, Yonatan},
  booktitle={Proceedings of the IEEE/CVF International Conference on Computer Vision},
  pages={7030--7038},
  year={2023}
}

@article{rassin2024linguistic,
  title={Linguistic Binding in Diffusion Models: Enhancing Attribute 
         Correspondence through Attention Map Alignment},
  author={Rassin, Royi and Hirsch, Eran and Glickman, Daniel and 
          Ravfogel, Shauli and Goldberg, Yoav and Chechik, Gal},
  journal={Advances in Neural Information Processing Systems},
  volume={36},
  year={2024}
}

@inproceedings{huang2023t2icompbench,
  title={T2I-CompBench: A Comprehensive Benchmark for Open-world 
         Compositional Text-to-Image Generation},
  author={Huang, Kaiyi and Sun, Kaiyue and Xie, Enze and Li, Zhenguo 
          and Liu, Xihui},
  booktitle={Advances in Neural Information Processing Systems},
  volume={36},
  year={2023}
}

@article{trusca2024object,
  title={Object-attribute binding in text-to-image generation: Evaluation and control},
  author={Trusca, Maria Mihaela and Nuyts, Wolf and Thomm, Jonathan and Honig, Robert and Hofmann, Thomas and Tuytelaars, Tinne and Moens, Marie-Francine},
  journal={arXiv preprint arXiv:2404.13766},
  year={2024}
}

@book{cover2006elements,
  title={Elements of Information Theory},
  author={Cover, Thomas M. and Thomas, Joy A.},
  edition={2},
  year={2006},
  publisher={Wiley-Interscience}
}

@article{fu2024commonsense,
  title={Commonsense-T2I challenge: Can text-to-image generation models understand commonsense?},
  author={Fu, Xingyu and He, Muyu and Lu, Yujie and Wang, William Yang and Roth, Dan},
  journal={arXiv preprint arXiv:2406.07546},
  year={2024}
}

@article{huberman2025image,
  title={Image Generation from Contextually-Contradictory Prompts},
  author={Huberman, Saar and Patashnik, Or and Dahary, Omer and Mokady, Ron and Cohen-Or, Daniel},
  journal={arXiv preprint arXiv:2506.01929},
  year={2025}
}

@article{yang2024position,
  title={Position: towards implicit prompt for text-to-image models},
  author={Yang, Yue and Lin, Yuqi and Liu, Hong and Shao, Wenqi and Chen, Runjian and Shang, Hailong and Wang, Yu and Qiao, Yu and Zhang, Kaipeng and Luo, Ping},
  journal={arXiv preprint arXiv:2403.02118},
  year={2024}
}

@inproceedings{parast2025ddb,
  title={DDB: Diffusion Driven Balancing to Address Spurious Correlations},
  author={Parast, Aryan Yazdan and Azam, Basim and Akhtar, Naveed},
  booktitle={Proceedings of the IEEE/CVF International Conference on Computer Vision},
  pages={17526--17535},
  year={2025}
}

@article{vice2026fairness,
  title={On the fairness, diversity and reliability of text-to-image generative models},
  author={Vice, Jordan and Akhtar, Naveed and Sigal, Leonid and Hartley, Richard and Mian, Ajmal},
  journal={Artificial Intelligence Review},
  volume={59},
  number={2},
  pages={57},
  year={2026},
  publisher={Springer}
}

@inproceedings{azam2025plug,
  title={Plug-and-Play Interpretable Responsible Text-to-Image Generation via Dual-Space Multi-facet Concept Control},
  author={Azam, Basim and Akhtar, Naveed},
  booktitle={Proceedings of the Computer Vision and Pattern Recognition Conference},
  pages={2976--2985},
  year={2025}
}

@inproceedings{seshadri2024bias,
  title={The bias amplification paradox in text-to-image generation},
  author={Seshadri, Preethi and Singh, Sameer and Elazar, Yanai},
  booktitle={Proceedings of the 2024 Conference of the North American Chapter of the Association for Computational Linguistics: Human Language Technologies (Volume 1: Long Papers)},
  pages={6367--6384},
  year={2024}
}

@inproceedings{carlini2023extracting,
  title={Extracting training data from diffusion models},
  author={Carlini, Nicolas and Hayes, Jamie and Nasr, Milad and Jagielski, Matthew and Sehwag, Vikash and Tramer, Florian and Balle, Borja and Ippolito, Daphne and Wallace, Eric},
  booktitle={32nd USENIX security symposium (USENIX Security 23)},
  pages={5253--5270},
  year={2023}
}

@inproceedings{kernel,
  title={Demystifying {MMD GANs}},
  author={Bi\'{n}kowski, Miko{\l}aj and Sutherland, Danica J. and Arbel, Michael and Gretton, Arthur},
  booktitle={International Conference on Learning Representations},
  year={2018}
}

@article{seo2025geometrical,
  title={Geometrical Properties of Text Token Embeddings for Strong Semantic Binding in Text-to-Image Generation},
  author={Seo, Hoigi and Bang, Junseo and Lee, Haechang and Lee, Joohoon and Lee, Byung Hyun and Chun, Se Young},
  journal={arXiv preprint arXiv:2503.23011},
  year={2025}
}

@inproceedings{phung2024grounded,
  title={Grounded text-to-image synthesis with attention refocusing},
  author={Phung, Quynh and Ge, Songwei and Huang, Jia-Bin},
  booktitle={Proceedings of the IEEE/CVF Conference on Computer Vision and Pattern Recognition},
  pages={7932--7942},
  year={2024}
}

@article{hu2024ella,
  title={Ella: Equip diffusion models with llm for enhanced semantic alignment},
  author={Hu, Xiwei and Wang, Rui and Fang, Yixiao and Fu, Bin and Cheng, Pei and Yu, Gang},
  journal={arXiv preprint arXiv:2403.05135},
  year={2024}
}

@inproceedings{sun2025dreamsync,
  title={Dreamsync: Aligning text-to-image generation with image understanding feedback},
  author={Sun, Jiao and Fu, Deqing and Hu, Yushi and Wang, Su and Rassin, Royi and Juan, Da-Cheng and Alon, Dana and Herrmann, Charles and Van Steenkiste, Sjoerd and Krishna, Ranjay and others},
  booktitle={Proceedings of the 2025 Conference of the Nations of the Americas Chapter of the Association for Computational Linguistics: Human Language Technologies (Volume 1: Long Papers)},
  pages={5920--5945},
  year={2025}
}

@article{he2024implicit,
  title={Implicit Priors Editing in Stable Diffusion via Targeted Token Adjustment},
  author={He, Feng and Zhang, Chao and Zhao, Zhixue},
  journal={arXiv preprint arXiv:2412.03400},
  year={2024}
}

@inproceedings{coco,
  title={Microsoft coco: Common objects in context},
  author={Lin, Tsung-Yi and Maire, Michael and Belongie, Serge and Hays, James and Perona, Pietro and Ramanan, Deva and Doll{\'a}r, Piotr and Zitnick, C Lawrence},
  booktitle={European conference on computer vision},
  pages={740--755},
  year={2014},
  organization={Springer}
}

@article{islam2026genmix,
  title={Genmix: effective data augmentation with generative diffusion model image editing},
  author={Islam, Khawar and Zaheer, Muhammad Zaigham and Mahmood, Arif and Nandakumar, Karthik and Akhtar, Naveed},
  journal={Expert Systems with Applications},
  pages={132273},
  year={2026},
  publisher={Elsevier}
}

@article{islam2025context,
  title={Context-guided responsible data augmentation with diffusion models},
  author={Islam, Khawar and Akhtar, Naveed},
  journal={arXiv preprint arXiv:2503.10687},
  year={2025}
}


%% file: supp.bib
@article{wu2025nep,
  title={Nep: Autoregressive image editing via next editing token prediction},
  author={Wu, Huimin and Ma, Xiaojian and Zhao, Haozhe and Zhao, Yanpeng and Li, Qing},
  journal={arXiv preprint arXiv:2508.06044},
  year={2025}
}

@inproceedings{sajnani2025geodiffuser,
  title={Geodiffuser: Geometry-based image editing with diffusion models},
  author={Sajnani, Rahul and Vanbaar, Jeroen and Min, Jie and Katyal, Kapil D and Sridhar, Srinath},
  booktitle={Proceedings of the Winter Conference on Applications of Computer Vision},
  pages={472--482},
  year={2025}
}

@inproceedings{rohrbach2018object,
  title={Object hallucination in image captioning},
  author={Rohrbach, Anna and Hendricks, Lisa Anne and Burns, Kaylee and Darrell, Trevor and Saenko, Kate},
  booktitle={Proceedings of the 2018 Conference on Empirical Methods in Natural Language Processing},
  pages={4035--4045},
  year={2018}
}

@inproceedings{cho2024davidsonian,
  title={Davidsonian scene graph: Improving reliability in fine-grained evaluation for text-to-image generation},
  author={Cho, Jaemin and Hu, Yushi and Baldridge, Jason and Garg, Roopal and Anderson, Peter and Krishna, Ranjay and Bansal, Mohit and Pont-Tuset, Jordi and Wang, Su},
  booktitle={International conference on learning representations},
  volume={2024},
  pages={15625--15645},
  year={2024}
}
